\documentclass[10pt,twocolumn,letterpaper]{article}

\usepackage{iccv}
\usepackage{times}
\usepackage{epsfig}
\usepackage{graphicx}
\usepackage{amsmath}
\usepackage{amssymb}
\usepackage{booktabs, makecell, tabularx}
\usepackage{multirow}
\usepackage{dsfont}
\usepackage{enumitem}
\usepackage{algorithm}
\usepackage{algpseudocode}
\usepackage{xspace}

\usepackage[pagebackref=true,breaklinks=true,colorlinks,bookmarks=false]{hyperref}

\iccvfinalcopy %

\def\iccvPaperID{****} %

\ificcvfinal\fi
\usepackage{color}
\usepackage[dvipsnames]{xcolor}
\usepackage{caption}

\newcommand{\reffig}[1]{Figure~\ref{fig:#1}}
\newcommand{\refsec}[1]{Section~\ref{sec:#1}}

\newcommand{\lblfig}[1]{\label{fig:#1}}

\newcommand{\myparagraph}[1]{\vspace{-2pt}\paragraph{#1}}
\newcommand{\ours}{pix2pix-zero\xspace}
\newcommand{\prompt}{c\xspace}

\begin{document}

\title{Zero-shot Image-to-Image Translation}

\author{Gaurav Parmar\textsuperscript{1}
\qquad
Krishna Kumar Singh\textsuperscript{2}
\qquad
Richard Zhang\textsuperscript{2}
\qquad \\
Yijun Li\textsuperscript{2}
\qquad
Jingwan Lu\textsuperscript{2}
\qquad
Jun-Yan Zhu\textsuperscript{1}\\
\textsuperscript{1}Carnegie Mellon University
\qquad
\textsuperscript{2}Adobe Research
}

\vspace{-6mm}
\twocolumn[{%
\renewcommand\twocolumn[1][]{#1}%
\maketitle

\begin{center}
    \centering
    \includegraphics[width=\linewidth]{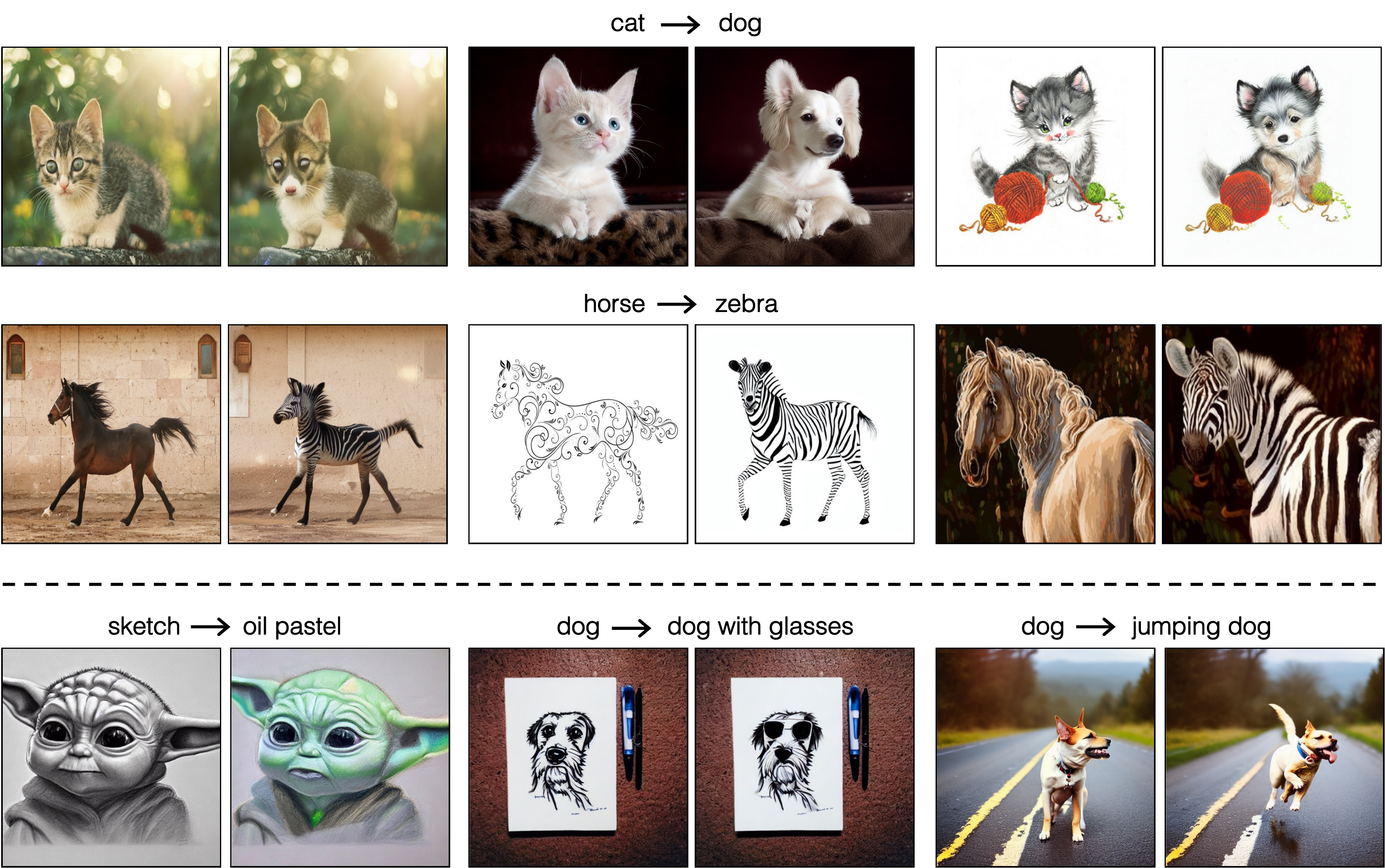}
    \vspace{-3mm}
    \captionof{figure}{We propose \texttt{pix2pix-zero},  a diffusion-based image-to-image translation method that allows users to specify the edit direction
    on-the-fly (e.g., cat $\rightarrow$ dog). We perform various translation tasks on both real (top 2 rows) and synthetic (bottom row) images, while preserving the structure of the input image. Our method requires \emph{neither} manual text prompting for each input image \emph{nor} costly fine-tuning for each task. }%
    \label{fig:teaser}
\end{center}
}]
 \maketitle
\ificcvfinal\thispagestyle{empty}\fi

\begin{abstract}
\vspace{-4mm}
Large-scale text-to-image generative models have shown their remarkable ability to synthesize diverse and high-quality images. However, it is still challenging to directly apply these models for editing real images for two reasons. First, it is hard for users to come up with a perfect text prompt that accurately describes every visual detail in the input image. 
Second, while existing models can introduce desirable changes in certain regions, they often dramatically alter the input content and introduce unexpected changes in unwanted regions. 
In this work, we propose \ours, an image-to-image translation method that can preserve the content of the original image without manual prompting. We first automatically discover editing directions that reflect desired edits in the text embedding space.
To preserve the general content structure after editing, we further propose cross-attention guidance, which aims to retain the cross-attention maps of the input image throughout the diffusion process. In addition, our method does not need additional training for these edits and can directly use the existing pre-trained text-to-image diffusion model.
We conduct extensive experiments and show that our method outperforms existing and concurrent works for both real and synthetic image editing.  
\end{abstract}

\section{Introduction}
\label{sec:intro}

Recent text-to-image diffusion models, such as DALL·E 2~\cite{ramesh2022hierarchical}, Imagen~\cite{saharia2022photorealistic} and Stable Diffusion~\cite{rombach2022high} generate diverse and realistic synthetic images with complex objects and scenes, displaying powerful compositional ability.
However, repurposing such models for editing \textit{real} images remains challenging.

First, images do not naturally come with text descriptions.
Specifying one is cumbersome and time-consuming, as a picture is worth the proverbial ``thousand words'', containing many texture details, lighting conditions, and shape subtleties that may not have corresponding words in the vocabulary. Second, even with initial and target text prompts (e.g., changing the word from \emph{cat} to \emph{dog}), existing text-to-image models tend to synthesize completely new content that fails to follow the layout, shape, and object pose of the input image. After all, editing the text prompt only tells us what we want to \textit{change}, but does not convey what we intend to \textit{preserve}. 
Finally, users may want to perform all kinds of edits on a diverse set of real images. %
So, we do not want to finetune a large model for each image and edit type due to its prohibitive costs. %

To overcome the above issues, we introduce \texttt{\ours}, a diffusion-based image-to-image translation approach that is \emph{training-free} and \emph{prompt-free}.  A user only needs to specify the edit direction in the form of source domain $\rightarrow$ target domain (e.g., cat$\rightarrow$ dog) on-the-fly, without manually creating text prompts for the input image. Our model can directly use pre-trained text-to-image diffusion models, without additional training for each edit type and image.

In this work, we make two key contributions: (1) \emph{An efficient, automatic editing direction discovery mechanism without input text prompting.} We automatically discover generic edit directions that work for a wide range of input images. %
Given an original word (e.g., cat) and an edited word (e.g., dog), we generate two groups of sentences containing the original and edited words separately. Then we compute the CLIP embedding direction between the two groups. %
As this editing direction is based on multiple sentences, it is more robust than just finding the direction only between the original and edited words. This step only takes about 5 seconds and can be pre-computed.  (2) \emph{Content preservation via cross-attention guidance.} Our observation is that the cross-attention map corresponds to the structure of the generated object. To preserve the original structure, we encourage the text-image cross-attention map to be consistent before and after translation. %
Hence, we apply the cross-attention guidance to enforce this consistency throughout the diffusion process. In Figure ~\ref{fig:teaser}, we show various editing results using our method while preserving the structure of input images. %
 
We further improve our results and inference speed with a suite of techniques: (1) Autocorrelation regularization: When applying inversion via DDIM~\cite{song2021ddim}, we observe that DDIM inversion is prone to make intermediate predicted noise less Gaussian, which reduces the edibility of an inverted image. Hence, we introduce an autocorrelation regularization to ensure noise to be close to Gaussian during inversion. (2) Conditional GAN distillation: Diffusion models are slow due to the multi-step inference of a costly diffusion process. To enable interactive editing, we distill the diffusion model to a fast conditional GAN model, given paired data of the original and edited images from the diffusion model, enabling real-time inference. %

We demonstrate our method on a wide range of image-to-image translation tasks, such as changing the foreground object (cat$\rightarrow$ dog), modifying the object (adding glasses to a cat image), and changing the style of the input (sketch $\rightarrow$ oil pastel), for both real images and synthetic images. Extensive experiments show that \ours outperforms existing and concurrent works~\cite{meng2021sdedit,hertz2022prompt} regarding photorealism and content preservation. %
Finally, we include an extensive ablation study on individual algorithmic components and discuss our method's limitations. 
See our website \url{https://pix2pixzero.github.io/} for additional results and the accompanying code.

\section{Related Work} 
\label{sec:related}
\myparagraph{Deep image editing with GANs.}
With generative modeling, image editing techniques have enabled users to express their goals in different ways (e.g., a slider, a spatial mask, or a natural language description). %
One line of work is to train conditional GANs that translate an input image from one domain to a target domain~\cite{isola2017image,sangkloy2016scribbler,zhu2017unpaired,choi2017stargan,wang2018pix2pixHD,huang2018multimodal,park2019semantic,liu2017unsupervised,baek2021rethinking}, which often requires task-specific model training. 
Another category of editing approaches is manipulating the latent space of GANs via image inverting the image and discovering the editing direction \cite{zhu2016generative, huh2020transforming, richardson2020encoding, zhu2020indomain, wulff2020improving,bau2021paint}. They first project the target image to the latent space of a pretrained GAN model and then edit the image by manipulating the latent code along directions corresponding to disentangled attributes. Numerous prior works propose to finetune the GAN model to better match the input image\cite{bau2019semantic,pan2021exploiting,roich2021pivotal}, explore different latent spaces \cite{wu2020stylespace, abdal2019image2stylegan, abdal2020image2stylegan++}, invert into multiple layers \cite{gu2020image, parmar2022sam}, and utilize latent edit directions \cite{harkonen2020ganspace, shen2020interfacegan, patashnik2021styleclip, abdal2022clip2stylegan}. While these methods are successful on single-category curated datasets, they struggle to obtain a high-quality inversion on more complex images. %

\myparagraph{Text-to-Image models.} %
Recently, large-scale text-to-image models have dramatically improved the image quality and diversity by training on an internet-scale text-image datasets~\cite{saharia2022photorealistic,ramesh2022hierarchical,ramesh2021zero,yu2022scaling,ding2021cogview,gafni2022make}.
However, they provide limited control over the generation process outside the text input. 
Editing real images by changing words in the input sentence is not reliable as it often changes too much of the image in unexpected ways. Some methods~\cite{nichol2021glide,avrahami2022blended} use additional masks to constrain where edits are applied. Unlike these approaches, our method retains the input structure without any spatial mask. 
Other recent and concurrent works (e.g., Palette~\cite{saharia2022palette}, InstructPix2Pix~\cite{brooks2022instructpix2pix}, PITI~\cite{wang2022pretraining})  learn conditional diffusion models tailored for image-to-image translation tasks. In contrast, we use the pre-trained Stable Diffusion models, without additional training.

\myparagraph{Image editing with diffusion models.}
Several recent works have adopted diffusion models for image editing. SDEdit~\cite{meng2021sdedit} performs editing by first adding noise to the input image together with a user editing guide, and then denoising it to increase its realism.
It is later used with text-to-image models such as GLIDE~\cite{nichol2021glide} and Stable Diffusion models~\cite{rombach2022high} to perform text-based image inpainting and editing. Other methods~\cite{choi2021ilvr,song2020score} propose to modify the diffusion process by incorporating conditioning user inputs but have been only applied to single-category models. %

 Two concurrent works, Imagic~\cite{kawar2022imagic} and prompt-to-prompt~\cite{hertz2022prompt}, also attempt structure-preserving editing via pretrained text-to-image diffusion models. Imagic~\cite{kawar2022imagic} demonstrates great editing results but requires finetuning the entire model for each image. %
 Prompt-to-prompt~\cite{hertz2022prompt} does not require finetuning and uses the cross-attention map of the original image with values corresponding to edited text to retain structure, with a main focus on synthetic image editing. Our work differs in three ways. First, our method requires no text prompting for the input image. Second, our approach is more robust as we do not directly use the cross-attention map of the original text, which may be incompatible with edited text. Our guidance-based method ensures the cross-attention map of edited images remains close but still has the flexibility to change according to edited text. 
 Third, our method is tailored for real images, while still being effective for synthetic ones. 
We show that our method outperforms SDEdit and prompt-to-prompt regarding image quality and content preservation.%

\section{Method} \label{sec:method}

\begin{figure}[t]
    \centering
    \includegraphics[width=\linewidth]{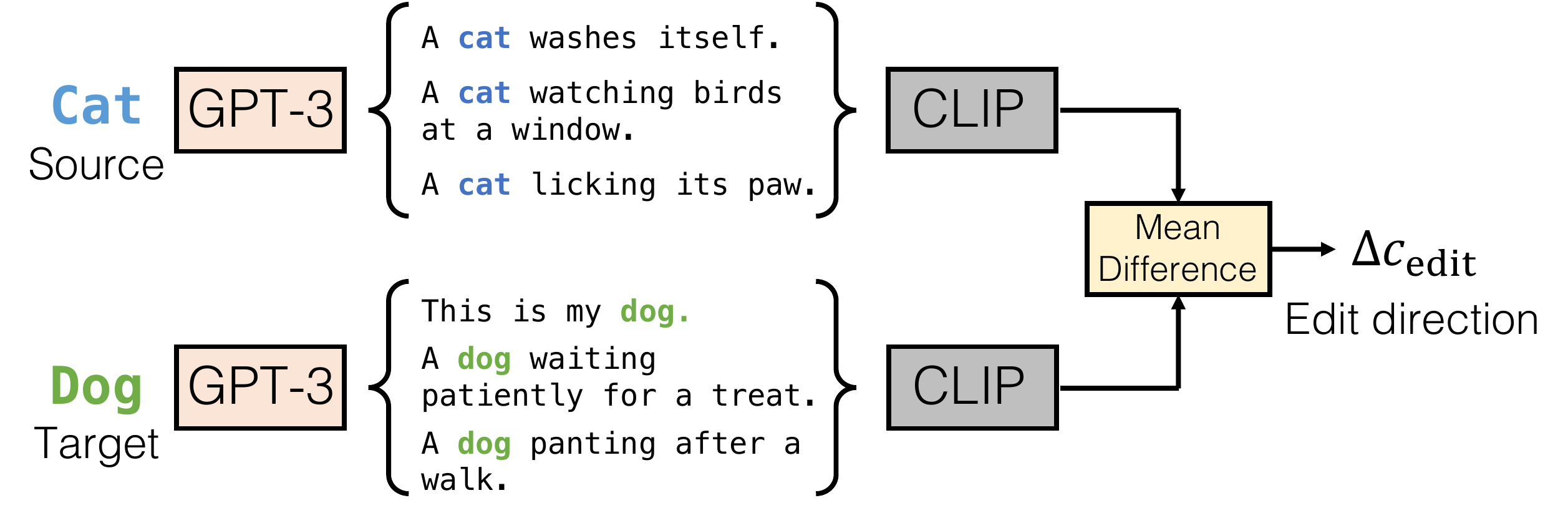}
    \vspace{-4mm}
    \caption{
    \textbf{Discovering edit directions}. Given the source and target text (e.g., cat and dog), we generate a large bank of diverse sentences using GPT-3. We compute their CLIP embeddings and take the mean difference to obtain edit direction $\Delta c_\text{edit}$.
    }
    \label{fig:method_text_dir}
\vspace{-1.5em}
\end{figure}
\begin{figure*}[t!]
    \centering
    \includegraphics[width=\linewidth]{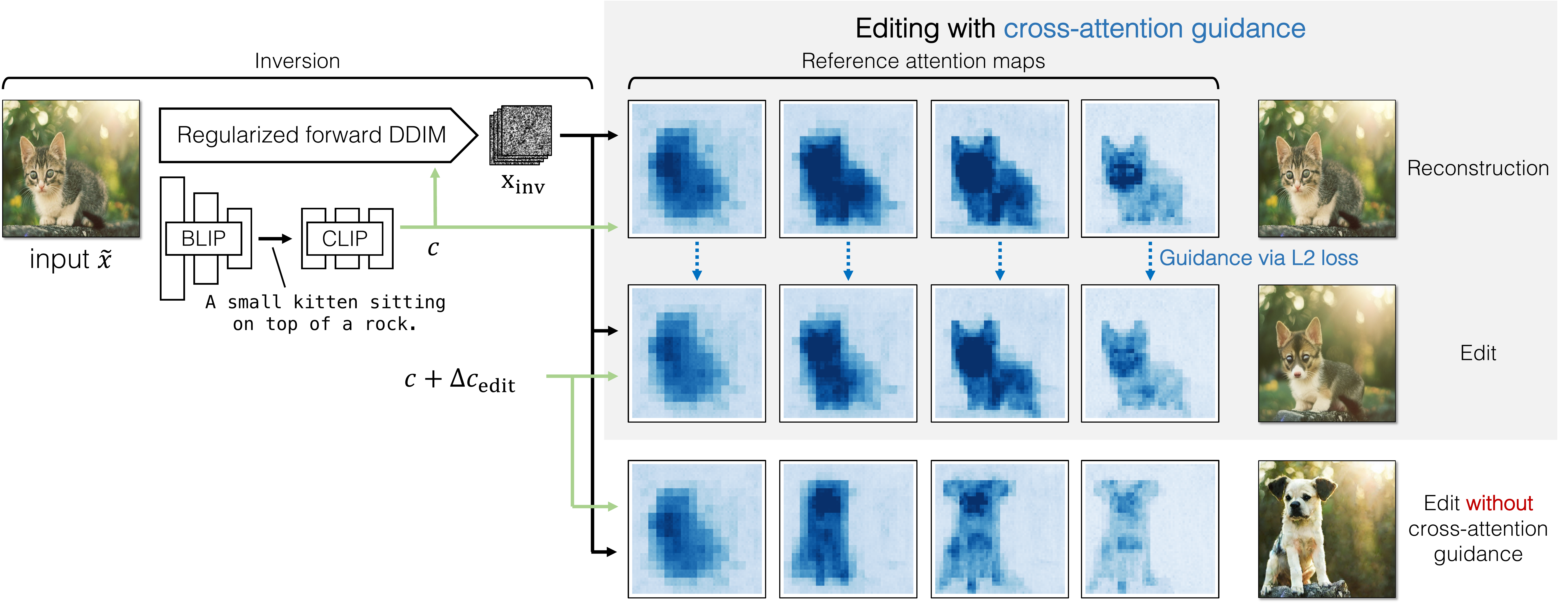}
    \caption{{\bf Overview of the \ours method}, illustrated by a cat $\rightarrow$ dog editing example. %
    First, we apply our regularized DDIM inversion to obtain an inverted noise map. This is guided by text embedding $\prompt$, automatically computed using image captioning network BLIP~\cite{li2022blip} and the CLIP text embedding model. Then, we denoise with the original text embedding to obtain cross-attention maps, serving as a reference for the input image structure (top row). Next, we denoise with the \textit{edited} text embedding, $\prompt + \Delta c_\text{edit}$, using a loss to encourage the cross-attention maps to match the reference cross-attention maps (2nd row). This ensures the structure of edited images does not change dramatically, compared to the original image. %
    Denoising \emph{without} cross-attention guidance is shown in the 3$^\text{rd}$ row, resulting in large deviations in structure.
      }
    \lblfig{pipeline}
\end{figure*}

We propose to edit an input image along an \textit{edit direction} (e.g., cat $\rightarrow$ dog).
We first invert the input $\Tilde{x}$ in a deterministic manner to the corresponding noise map in \refsec{inversion}. In \refsec{direction}, we present a method for automatically discovering and pre-computing edit directions in text embedding space. Applying the edit direction naively often results in unwanted changes in image content. To address this issue, we propose cross-attention guidance that guides the diffusion sampling process and helps retain the input image's structure (\refsec{cross_attention}). Note that our method is applicable to different text-to-image models but for this paper we use Stable Diffusion~\cite{rombach2022high} which encodes input image $\Tilde{x}\in \mathds{R}^{X\times X\times 3}$ to latent code $x_0 \in \mathds{R}^{S\times S\times 4}$. In our experiments, $X=512$ is the image size, and $S=64$ is the downsampled latent size. The inversion and editing described in this section happen in the latent space. To invert a text-conditioned model, we generate an initial text prompt $\prompt$ using BLIP~\cite{li2022blip} to describe the input image 
$\Tilde{x}$.

\subsection{Inverting Real Images} \label{sec:inversion}

\myparagraph{Deterministic inversion.}
Inversion entails finding a noise map $x_\text{inv}$ that reconstructs the input latent code $x_0$ upon sampling. In DDPM~\cite{ho2020ddpm}, this corresponds to the fixed forward noising process, followed by de-noising with the reverse process. However, both the forward and reverse processes of DDPM are stochastic and do not result in a faithful reconstruction. 
Instead, we adopt the deterministic DDIM~\cite{song2021ddim} reverse process, as shown below: %
\begin{equation}
    x_{t+1} = \sqrt{\bar{\alpha}_{t+1}}f_\theta(x_t,t,c) + \sqrt{1-\bar{\alpha}_{t+1}} \epsilon_\theta(x_t,t,\prompt),
\end{equation}

\noindent where $x_{t}$ is noised latent code at timestep $t$, $\epsilon_\theta(x_t,t,\prompt)$ is a UNet-based denoiser that predicts added noise in $x_{t}$ conditional on timestep $t$ and encoded text features $\prompt$, $\bar{\alpha}_{t+1}$ is noise scaling factor as defined in DDIM~\cite{song2021ddim}, and $f_\theta(x_t,t,c)$ predicts the final denoised latent code $x_0$.

\begin{equation}
    f_\theta(x_t,t,c) = \frac{x_t - \sqrt{1-\bar{\alpha}_t} \epsilon_\theta(x_t,t,\prompt) }{\sqrt{\bar{\alpha}_t}}
\end{equation}

\noindent We gradually add noise to initial latent code $x_0$ using DDIM process and at the end of inversion, the final noised latent code $x_{T}$ is assigned as $x_\text{inv}$.

\myparagraph{Noise regularization.}
The inverted noise maps generated by DDIM inversion $\epsilon_\theta(z_t, t, \prompt) \in \mathds{R}^{S\times S\times 4}$ often do not follow the statistical properties of uncorrelated, Gaussian white noise, causing %
poor editability. A Gaussian white noise map should have (1) no correlation between any pair of random locations and (2) zero-mean, unit-variance at each spatial location, which would be reflected in its auto-correlation function being a Kronecker delta function~\cite{gubner2006probability}. Following this, we guide the inversion process with an auto-correlation objective, comprised of a pairwise term $\mathcal{L}_{\text{pair}}$ and a KL divergence term $\mathcal{L}_{\text{KL}}$ at individual pixel locations.

As densely sampling all pairs of locations is costly, we follow~\cite{karras2020analyzing} and form a pyramid, where the initial noise level $\eta^0\in \mathds{R}^{64\times 64\times 4}$ is the predicted noise map $\epsilon_\theta$, and each subsequent noise map is average pooled with a $2\times2$ neighborhood (and multiplied by 2, to preserve the expected variance). We stop at feature size $8\times 8$, creating 4 noise maps to form set $\{\eta^0, \eta^1, \eta^2, \eta^3\}$.

The pairwise regularization at pyramid level $p$ is the sum of squares of the auto-correlation coefficients at possible $\delta$ offsets, normalized over noise map sizes $S_p$.

\begin{equation}
\begin{aligned}
    \mathcal{L}_{\text{pair}} = \sum_p \frac{1}{S_p^2} \sum_{\delta=1}^{S_{p}-1} \sum_{x,y,c} \eta^p_{x,y,c} \left(\eta^p_{x-\delta, y,c}
    + \eta^p_{x, y-\delta,c} \right),
\end{aligned}
\end{equation}

\noindent where $\eta^p_{x,y,c}\in \mathds{R}$ indexes into a spatial location, using circular indexing, and channel. Note that Karras et al.~\cite{karras2020analyzing} previously explored using an autocorrelation regularizer for GAN inversion into a noise map. We introduce a few changes to the autocorrelation idea to boost its performance in the diffusion context: we randomly sample a shift at each iteration, rather than only using $\delta=1$ as in~\cite{karras2020analyzing}, enabling us to propagate long-range information more efficiently. We hypothesize that in the diffusion context, it is important for each time step to be well-regularized, as relying on multiple iterations to propagate long-range connections causes intermediate time steps to fall out of distribution.

In addition, we find that enforcing the zero-mean unit-variance criteria strictly via normalization~\cite{karras2020analyzing} leads to divergence during the denoising process. Instead, we formulate this softly as a loss $\mathcal{L}_{\text{KL}}$, as used in variational autoencoders~\cite{kingma2013auto}.
This enables us to softly balance between the two losses. Our final autocorrelation regularization is $\mathcal{L}_{\text{auto}} = \mathcal{L}_{\text{pair}} + \lambda \mathcal{L}_{\text{KL}}$, where $\lambda$ balances the two terms.

\subsection{Discovering Edit Directions}\label{sec:direction}

Recent large generative models allow users to control the image synthesis by specifying a sentence that describes the output image. We instead want to provide the users with an interface where they only need to provide the desired \textit{change} from the source domain to the target domain (e.g., cat $\rightarrow$ dog).

\begin{figure*}
    \centering
    \includegraphics[width=\linewidth]{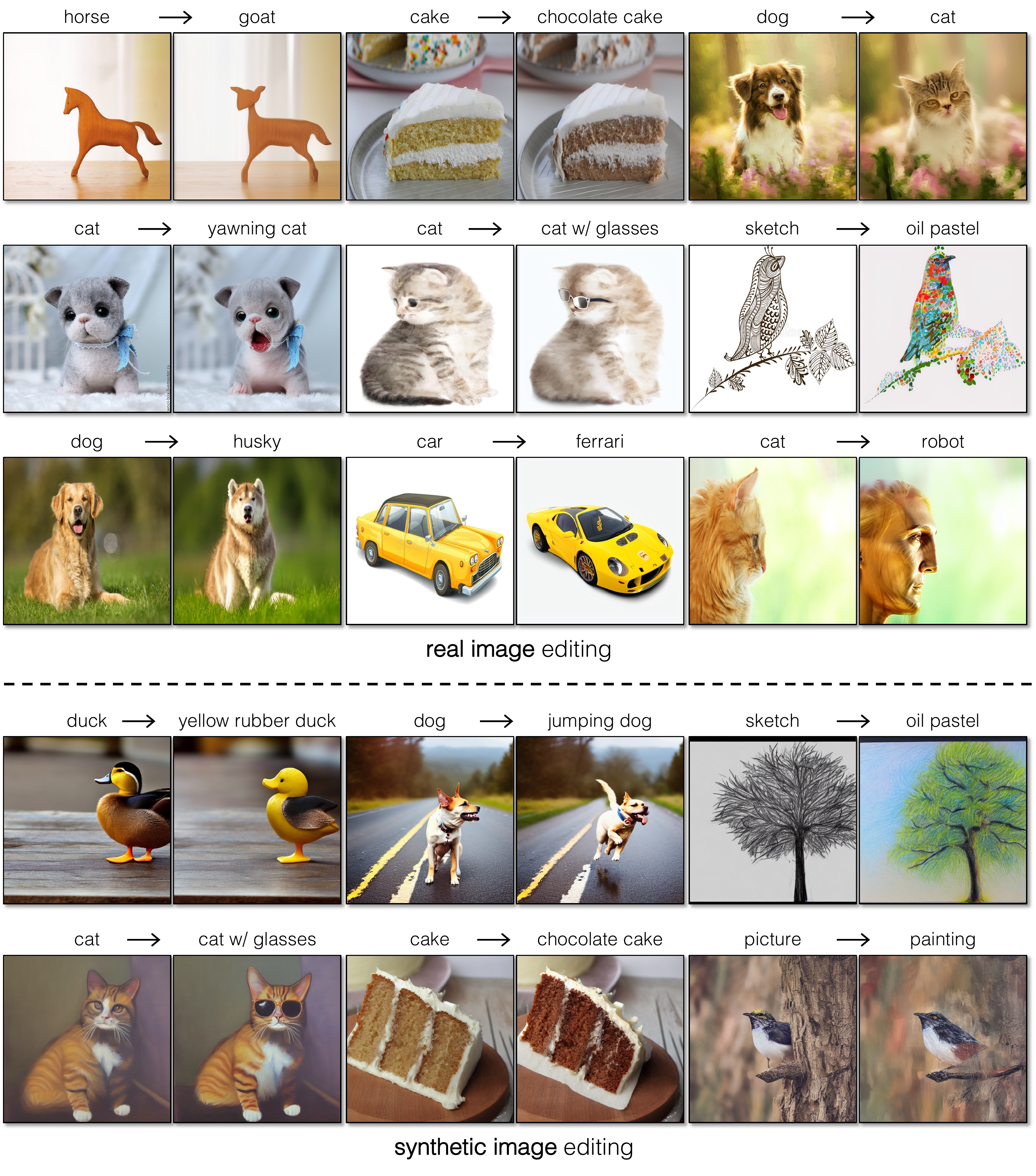}
    \vspace{-10pt}
    \caption{{\bf Examples of \ours results} on real (top) and synthetic images (bottom). For each image pair, we show the image before and after the edit. Note that the edit direction is generated from words alone (no prompts required).
    We are able to apply the edits while preserving the structure successfully.
    }
     \label{fig:text_directions}
\end{figure*}

We automatically compute the corresponding text embedding direction vector $\Delta{c_\text{edit}}$ from the source to the target, as illustrated in Figure~\ref{fig:method_text_dir}. We generate a large bank of diverse sentences for both source $s$ and the target $t$, either using an off-the-shelf sentence generator like GPT-3~\cite{brown2020language} or by using predefined prompts around source and target. We then compute the mean difference between CLIP embedding~\cite{radford2021learning} of the sentences. Edited images can be generated by adding the direction to the text prompt embedding. 
\reffig{text_directions} shows the result of several edits, with directions computed using this approach. We find text direction using multiple sentences more robust than a single word and demonstrate this in Section~\ref{sec:experiments}. %
This method of computing edit directions only takes about 5 seconds and only needs to be pre-computed once. 
Next, we incorporate the edit directions into our image-to-image translation method. %

\subsection{Editing via Cross-Attention Guidance}
\label{sec:cross_attention}
Recent large-scale diffusion models \cite{rombach2022ldm,saharia2022photorealistic,ramesh2022hierarchical} incorporate conditioning by augmenting the denoising network $\epsilon_\theta$ with the cross-attention layer \cite{bahdanau2014neural,vaswani2017attention}. 
We use the open-source Stable Diffusion model, built on latent diffusion Models (LDM)~\cite{rombach2022high}.
The model produces text embedding $\prompt$ with the CLIP \cite{radford2021learning} text encoder. Next,  to condition the generation on text, the model computes cross-attention between encoded text and intermediate features of the denoiser $\epsilon_\theta$: 
\begin{equation}
\begin{split}
    \text{Attention}&(Q,K,V) = M \cdot V, \\
    &\text{where } M=\text{Softmax} \left( \frac{QK^T}{\sqrt{d}} \right).
\end{split}
\end{equation}

\noindent Query $Q = W_Q \varphi(x_t)$, key $K = W_K \prompt$, and value $V = W_V \prompt$ are computed with the learnt projections $W_Q$, $W_K$, $W_V$ applied on intermediate spatial features $\varphi(x_t)$ of the denoising UNet $\epsilon_\theta$ and the text embedding $\prompt$, and $d$ is the dimension of projected keys and queries.
Of particular interest is the \textit{cross-attention map} $M$, which is observed to have a tight relation with the structure of the image~\cite{hertz2022prompt}.
Individual entries of the mask $M_{i,j}$ represent the contribution of the $j^\text{th}$-th text token towards the $i^\text{th}$ spatial location. Also, the cross-attention mask is specific to a timestep, and we get different attention mask $M_{t}$ for each timestep $t$. %

To apply an edit, the naive way would be to apply our pre-computed edit direction $\Delta c_\text{edit}$ to $\prompt$, and use $c_{\text{edit}} = \prompt + \Delta c_\text{edit}$ for the sampling process to generate $x_\text{edit}$.
This approach succeeds in changing the image according to the edit but fails to preserve the structure of the input image. As seen in the bottom row of Figure~\ref{fig:pipeline}, the deviation of the cross-attention maps during the sampling process results in deviation in the structure of the image.
As such, we propose a new \textit{cross-attention guidance} to encourage consistency in the cross-attention maps.

We follow a two-step process, as described in Algorithm \ref{alg:real_image_editing} and illustrated in Figure~\ref{fig:pipeline}. First, we reconstruct the image without applying the edit direction, just using the input text $c$ to obtain reference cross-attention maps $M_{t}^{\text{ref}}$ for each timestep $t$. These cross-attention maps correspond to the original image's structure e, which we aim to preserve. Next, we apply the edit direction by using $c_{edit}$ to generate cross-attention maps $M_{t}^{\text{edit}}$. We then take a gradient step with $x_t$ towards matching the reference $M_{t}^{\text{ref}}$, reducing the cross-attention loss $\mathcal{L}_{\text{xa}}$ below.

\begin{equation}
\mathcal{L}_{\text{xa}} = || M_{t}^{\text{edit}}-M_{t}^{\text{ref}} ||_2.
\end{equation}

\noindent This loss encourages $M_{t}^{\text{edit}}$ to not deviate from $M_t^{\text{ref}}$, applying the edit while retaining the original structure.

\begin{algorithm}[t!]
\caption{pix2pix-zero algorithm}
\label{alg:real_image_editing}
\begin{algorithmic}

\State \textbf{Input:} $x_T$ (\text{same as}  $x_{\text{inv}}$):  noise-regularized DDIM 
 \\
   \text{inversion of latent code corresponding to} $\Tilde{x}$ 
  \\ 
  \prompt: \text{input text features, } 
   \; $\Delta c_{\text{edit}}$: \text{edit direction} 
   \\
  $\lambda_{\text{xa}}$: \text{cross-attention guidance weight}
\\
 \State {\bf Output:} $x_0$ (\text{final edited latent code})

\\ \\
 $\triangleright$ Compute reference cross-attention maps

\For{$t = T...1$}
    \State $\hat{\epsilon}, M_{t}^{\text{ref}} \gets  \epsilon_\theta(x_t, t, \prompt)$
    \State $x_{t-1}=\Call{Update}{x_t, \hat{\epsilon}, t}$
    
\EndFor

\\
\State $\triangleright$ Edit with cross-attention guidance
\State $c_{\text{edit}} = \prompt + \Delta c_{\text{edit}}$
\For{$t = T...1$}

    \State $\_\_, M_{t}^{\text{edit}} \gets \epsilon_\theta(x_t, t, c_{\text{edit}})$
    \State $\Delta x_t = \nabla_{x_t}(|| M_{t}^{\text{edit}}-M_{t}^{\text{ref}} ||_2)$
    \State $\hat{\epsilon}, \_\_ \gets \epsilon_\theta \big(x_t - \lambda_{\text{xa}} \Delta x_t, t, c_{\text{edit}} \big)$
    \State $x_{t-1}=\Call{Update}{x_t, \hat{\epsilon}, t}$
    
\EndFor

\\
\State $\triangleright$ Update current state $x_t$ with noise prediction $\hat{\epsilon}$
\Function{Update}{$x_t, \hat{\epsilon}, t$}

    \State $x_{t-1} = \sqrt{\alpha_{t-1}} \; \frac{x_t - \sqrt{1-\alpha_t}\hat{\epsilon}}{\sqrt{\alpha_t}}+ \sqrt{1-\alpha_{t-1}}\hat{\epsilon}$
    \State \Return $x_{t-1}$
\EndFunction

\end{algorithmic}
\end{algorithm}

\begin{figure*}
    
    \centering
    \includegraphics[width=\linewidth]{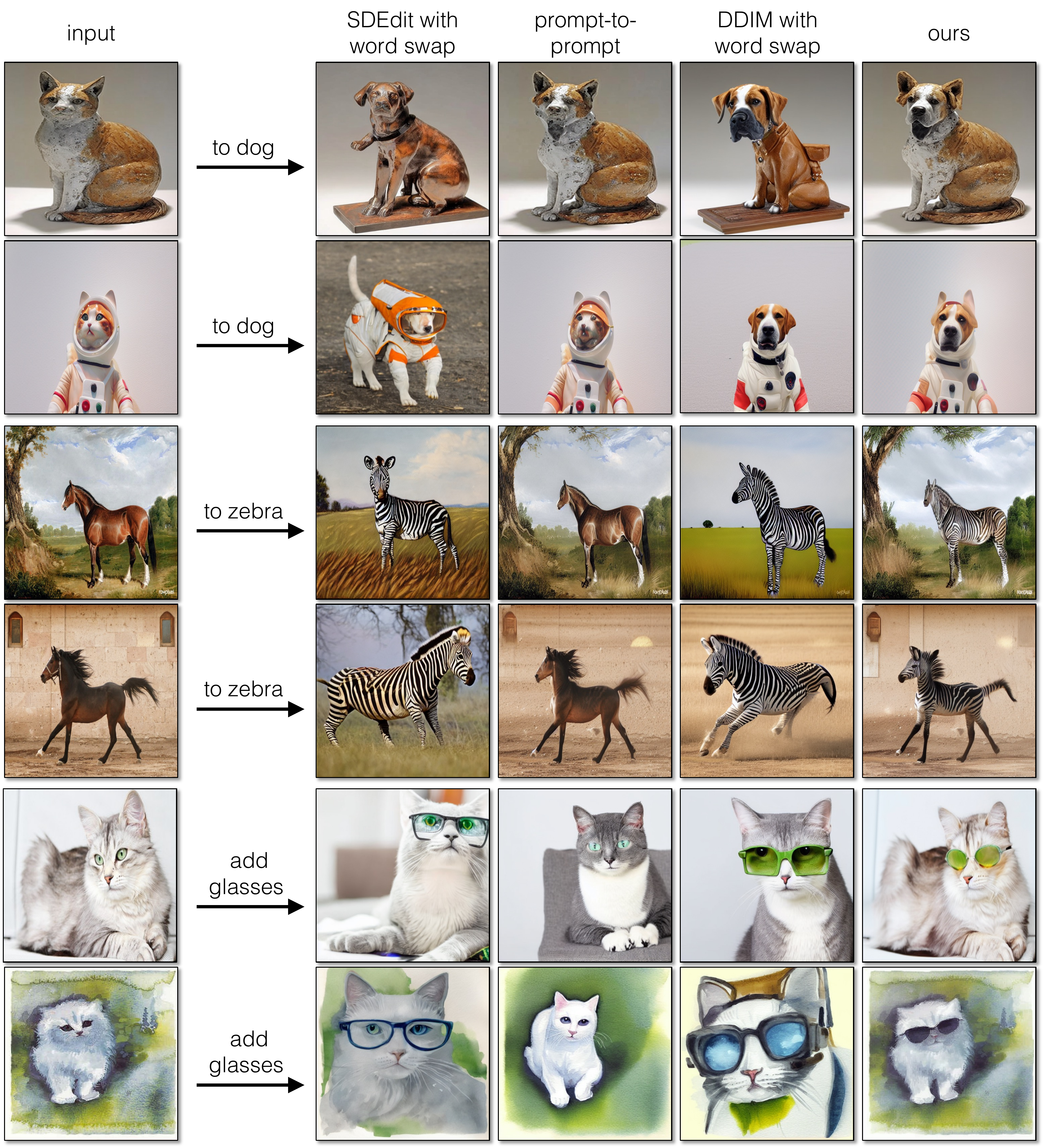}
    \caption{{\bf Comparisons with different baselines for real images}. We observe the SDEdit~\cite{meng2021sdedit} and DDIM~\cite{song2021ddim} + word swap methods show deviation in structure, while prompt-to-prompt~\cite{hertz2022prompt} struggles to perform the edit. Our method, as shown in the last column, successfully applies the edit, while preserving the structure of the input image.}
    \label{fig:cmp_baselines}
    
\end{figure*}
\section{Experiments} \label{sec:experiments}
Our image-to-image translation method can be used to edit real images and control the structure of synthetic images. Next, we demonstrate our method in various experiments using Stable Diffusion v1.4~\cite{stablediffusion}.  %

\subsection{Evaluation.} \label{sec:eval}

\myparagraph{Tasks.} We perform quantitative evaluations using four image-to-image translation tasks: (1) translating cats to dogs (cat $\rightarrow$ dog), (2) translating horses to zebras (horse $\rightarrow$ zebra), (3) starting with cat input images and adding glasses (cat $\rightarrow$ cat w/ glasses), (4) converting hand drawn sketches to oil pastel paintings (sketch $\rightarrow$ oil pastel). All input images are taken from LAION 5B dataset. See Appendix~\ref{sec:supp_details} for more details. These cover a large variety of edits, including changing the object (cat $\rightarrow$ dog, horse $\rightarrow$ zebra), modifying the object (cat $\rightarrow$ cat w/ glasses), and changing the global style (sketch $\rightarrow$ oil pastel).

\myparagraph{Metrics.}
For quantitative evaluations, we measure three criteria: (1) whether the edit was applied successfully, (2) whether the structure of the input image is retained in the edited image, and (3) if the background regions of the image stay unchanged. 
We measure the extent of the edit applied with CLIP Acc~\cite{hessel2021clipscore}, which calculates the percentage of instances where the edited image has a higher similarity to the target text, as measured by CLIP, than to the original source text. 
Subsequently, the structural consistency of the edited image is measured using Structure Dist~\cite{tumanyan2022splicing}. 
A lower score on Structure Dist means that the structure of the edited image is more similar to the input image. %
Lastly, to ensure that we retain the background after edits, we calculate the background LPIPS error (BG LPIPS). This is done by measuring the LPIPS distance between the background regions of the original and edited images. The background regions are identified using the object detector Detic \cite{zhou2022detecting}. A lower BG LPIPS score indicates that the background of the original image has been well preserved. 

The background error metric BG LPIPS is only applicable for specific editing tasks where only the foreground object needs to be altered (e.g. changing a cat to a dog, or a horse to a zebra). However, for editing tasks that involve changing the entire image (e.g. converting a sketch to an oil pastel), this metric is not relevant. 

\subsection{Qualitative Results} \label{sec:qual_result}
In Figure~\ref{fig:text_directions}, we show various edits applied by our approach on real (top) and synthetic images (bottom). For each result, we show pairs of images before and after editing. The edit direction is computed between the source and target, written on the top of each image pair. %
Our edit direction discovery method is capable of generating diverse edit directions, including changes in the type of object (e.g., from a dog to a cat or a horse to a goat), modifications of specific attributes of the object (e.g., adding sunglasses to a cat or making a cat yawn), and global style transformations of the image (e.g., from a sketch to an oil pastel or a photograph to a painting). The use of cross-attention guidance effectively preserves the structure of the original image.

\begin{table*}[t!]
    \centering
    \resizebox{\linewidth}{!}{
    \begin{tabular}{l ccc ccc cc cc}
        \toprule 
        \multirow{3}{*}{\textbf{Method}} 
        & \multicolumn{3}{c}{\textbf{cat $\rightarrow$ dog} }
        & \multicolumn{3}{c}{\textbf{horse $\rightarrow$ zebra} }
        & \multicolumn{2}{c}{\textbf{cat $\rightarrow$ cat w/ glasses} } 
        & \multicolumn{2}{c}{\textbf{sketch $\rightarrow$ oil pastel} } \\

        \cmidrule(lr){2-4} \cmidrule(lr){5-7} \cmidrule(lr){8-9} \cmidrule(lr){10-11}

        & \multirow{2}{*}{\shortstack[c]{CLIP-\\ Acc ($\uparrow$) }}  
        & \multirow{2}{*}{\shortstack[c]{BG\\ LPIPS ($\downarrow$) }} 
        & \multirow{2}{*}{\shortstack[c]{Structure \\ Dist ($\downarrow$) }} 
        
        & \multirow{2}{*}{\shortstack[c]{CLIP-\\ Acc ($\uparrow$) }}  
        & \multirow{2}{*}{\shortstack[c]{BG\\ LPIPS ($\downarrow$) }}
        & \multirow{2}{*}{\shortstack[c]{Structure \\ Dist ($\downarrow$) }}

        & \multirow{2}{*}{\shortstack[c]{CLIP-\\ Acc ($\uparrow$) }}  
        & \multirow{2}{*}{\shortstack[c]{Structure \\ Dist ($\downarrow$) }}

        & \multirow{2}{*}{\shortstack[c]{CLIP-\\ Acc ($\uparrow$) }}  
        & \multirow{2}{*}{\shortstack[c]{Structure \\ Dist ($\downarrow$) }}
        \\ \\
        \cmidrule(lr){1-11}

        SDEdit \cite{meng2021sdedit} + word swap 
        & 71.2\% & 0.327 & 0.081 
        & 92.2\% & 0.314 & 0.105
        & 34.0\% & 0.082
        & 21.2\% & 0.085 \\

        DDIM + word swap 
        & 72.0\% & 0.279 & 0.087
        & \textbf{94.0\% }& 0.283 & 0.123
        & 37.6\% & 0.085 
        & 32.4\% & 0.082 \\

        prompt-to-prompt \cite{hertz2022prompt} 
        & 66.0\% & 0.269 & 0.080 
        & 18.4\% & 0.261 & 0.095
        & 69.6\% & 0.081 
        & 10.8\% & 0.079\\
        
        \ours (ours) 
        & \textbf{92.4\%} & \textbf{0.182} & \textbf{0.044 }
        & 75.2\% & \textbf{0.194} & \textbf{0.066}
        & \textbf{71.2\%} & \textbf{0.028}
        & \textbf{75.2\%} & \textbf{0.052} \\
        \bottomrule 
    \end{tabular}
    }
    \caption{
    \textbf{Comparison to prior diffusion-based editing methods.} We compare our method to several prior diffusion-based image editing methods on four different tasks. The first two editing tasks  (cat $\rightarrow$ dog, horse $\rightarrow$ zebra) are evaluated with CLIP-Acc, BG LPIPS, and Structure Dist. These metrics assess the level of editing applied, the preservation of the background, and changes in the image structure changes, respectively. The other two tasks (cat $\rightarrow$ cat w/ glasses,  sketch $\rightarrow$ oil pastel) only use CLIP Acc and Structure Dist, as the background reconstruction is not relevant for these editing tasks. 
    Our method achieves the highest CLIP classification accuracy while retaining the details from the input image, as shown through a low background LPIPS score and low structure distance. 
    }
    \label{tab:cmp_baselines}
\end{table*}

\begin{figure*}[t!]
    \centering    \includegraphics[width=\linewidth]{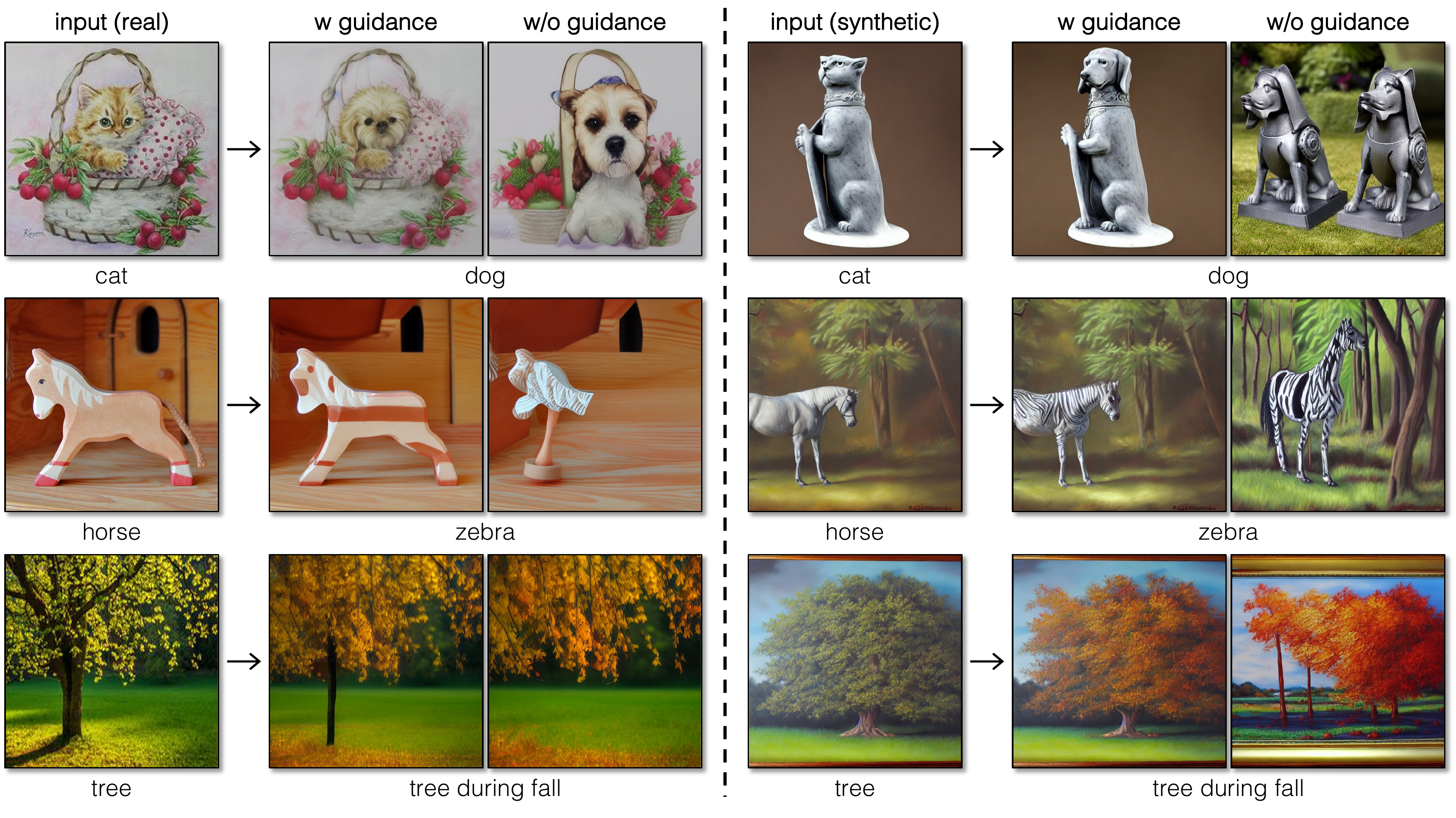}
    \caption{{\bf Effectiveness of cross-attention guidance on structure preservation}. We show the editing results for both real (left) and synthetic (right) images. With cross-attention guidance, the structure is well-preserved for objects. 
    }
    \label{fig:xa_comp}
\end{figure*}
\begin{figure}[t!]
    \centering
    \includegraphics[width=1.0\linewidth]{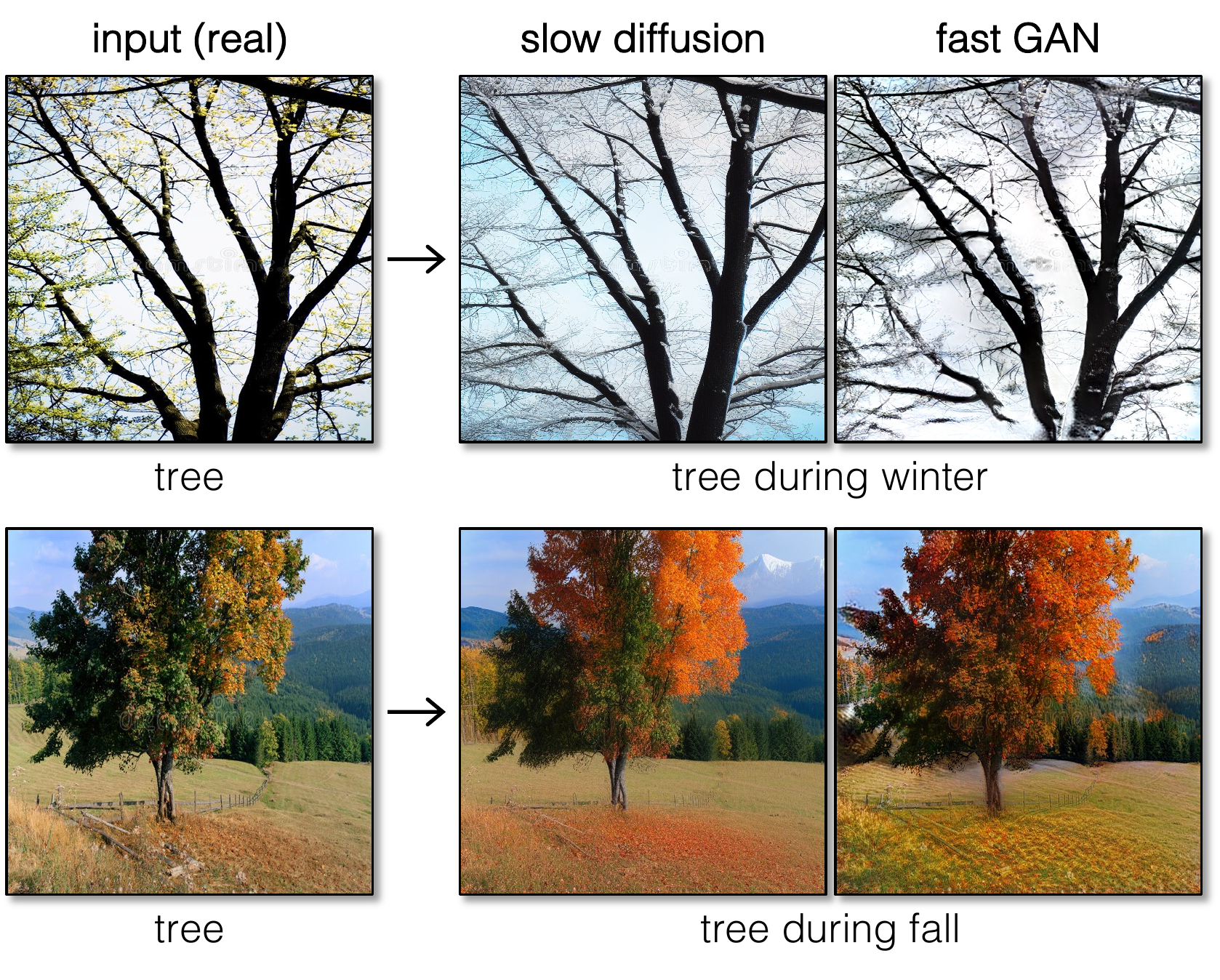}
    \vspace{-8mm}
    \caption{
    \textbf{Model acceleration with conditional GANs.} We show the results of the original diffusion-based model and conditional GANs for two tree editing tasks. The distilled GAN-based model achieves comparable results with a $\sim$ 3,800 times speedup. }
    \label{fig:distillation}
\end{figure}

\begin{table*}[t!]
    \centering
    \resizebox{\linewidth}{!}{
    \begin{tabular}{lll ccc ccc cc cc}
        \toprule 
        \multirow{3}{*}{\textbf{}} 
        & \multirow{3}{*}{\textbf{Inversion}} 
        & \multirow{3}{*}{\textbf{Edit}} 
        
        & \multicolumn{3}{c}{\textbf{cat $\rightarrow$ dog} }
        & \multicolumn{3}{c}{\textbf{horse $\rightarrow$ zebra} }
        & \multicolumn{2}{c}{\textbf{cat $\rightarrow$ cat w/ glasses} } 
        & \multicolumn{2}{c}{\textbf{sketch $\rightarrow$ oil pastel} } \\

        \cmidrule(lr){4-6} \cmidrule(lr){7-9} \cmidrule(lr){10-11} \cmidrule(lr){12-13}

        & & & \multirow{2}{*}{\shortstack[c]{CLIP-\\ Acc ($\uparrow$) }}  
        & \multirow{2}{*}{\shortstack[c]{BG\\ LPIPS ($\downarrow$) }} 
        & \multirow{2}{*}{\shortstack[c]{Structure \\ Dist ($\downarrow$) }} 
        
        & \multirow{2}{*}{\shortstack[c]{CLIP-\\ Acc ($\uparrow$) }}  
        & \multirow{2}{*}{\shortstack[c]{BG\\ LPIPS ($\downarrow$) }}
        & \multirow{2}{*}{\shortstack[c]{Structure \\ Dist ($\downarrow$) }}

        & \multirow{2}{*}{\shortstack[c]{CLIP-\\ Acc ($\uparrow$) }}  
        & \multirow{2}{*}{\shortstack[c]{Structure \\ Dist ($\downarrow$) }}

        & \multirow{2}{*}{\shortstack[c]{CLIP-\\ Acc ($\uparrow$) }}  
        & \multirow{2}{*}{\shortstack[c]{Structure \\ Dist ($\downarrow$) }}
        \\ \\
        \cmidrule(lr){1-13}

        \textbf{A} & DDPM & word swap 
        & 71.6\% & 0.392 & 0.126
        & 93.2\% & 0.389 & 0.167
        & 35.2\% & 0.122
        & 55.2\% & 0.114 \\
        
        \textbf{B} & DDIM & word swap 
        & 72.0\% & 0.279 & 0.087
        & 94.0\% & 0.283 & 0.123
        & 37.6\% & 0.085
        & 32.4\% & 0.082 \\

        \textbf{C} & DDIM  w/ $\mathcal{L}_{\text{auto}}$ & word swap 
        & 72.4\% & 0.283 & 0.089
        & 94.0\% & 0.281 & 0.122
        & 38.0\% & 0.087 
        & 35.2\% & 0.082 \\
        
        \textbf{D} & DDIM w/ $\mathcal{L}_{\text{auto}}$ & sentence directions 
        & \textbf{100.0\%} & 0.274 & 0.095
        & \textbf{97.6\%} & 0.290 & 0.130
        & 20.8\% & 0.103
        & \textbf{88.4\%} & 0.087 \\

        \textbf{E (ours)} & DDIM w/ $\mathcal{L}_{\text{auto}}$ & sentence directions w/ $\mathcal{L}_{\text{xa}}$ 
        & 92.4\% & \textbf{0.182} & \textbf{0.044}
        & 75.2\% & \textbf{0.194} & \textbf{0.066}
        & \textbf{71.2\%} & \textbf{0.028}
        & 75.2\% & \textbf{0.052} \\
       
        \bottomrule 
    \end{tabular}
    }
    \caption{
    \textbf{Ablation study.} We conduct an ablation study where we add different components of our method one at a time and observe the effects. We start with config A, which uses a naive stochastic DDPM noising process for inversion and word swap for applying the edit. This configuration does not retain the structure or the background of the input image. Config B, instead, uses deterministic DDIM inversion and results in the improvement of the structure and background preservation. Config C and D show that both regularized inversion ($\mathcal{L}_{\text{auto}}$) and sentence directions improve the editing ability. Config E, our final method, shows that using cross attention guidance $\mathcal{L}_{\text{xa}}$ improves the background and structure preservation.  }
    \label{tab:cmp_all_ablations}
\end{table*}
\subsection{Comparisons} \label{sec:comparison}
In this section, we compare our approach to some previous and concurrent diffusion-based image editing methods. For a fair comparison, all the approaches use the Stable Diffusion~\cite{stablediffusion} with the same number of sampling steps and the same classifier-free guidance for all methods. 
We compare against three baselines:

 {\bf 1) SDEdit~\cite{meng2021sdedit} + word swap:} this method first stochastically adds noise to an intermediate timestep and subsequently denoises with the new text prompt, where the source word is swapped with the target word.
 {\bf 2) Prompt-to-prompt~\cite{hertz2022prompt} (concurrent work):} we use the officially released code. The method swaps the source word with the target and uses the original cross-attention map as a \textit{hard} constraint. {\bf 3) DDIM + word swap:}  we invert with  the deterministic forward DDIM process and perform DDIM sampling with an edited prompt generated by swapping the source word with the target. %

In Figure~\ref{fig:cmp_baselines}, we compare our approach with the baselines. Both the SDEdit and DDIM + word swap methods struggle to retain the structure of the input image, as they do not use the cross-attention map of the original image. Prompt-to-prompt retains the cross-attention map of the original image as a hard constraint, thus the structure. However, this comes at the cost of not performing the desired edit. 
In contrast, our approach utilizes the original cross-attention map as soft guidance, implemented as a loss function, allowing for flexibility in the edited cross-attention map to adapt to the chosen edit direction. 
As a result, we can perform the edit while preserving the structure of the input image. 

In Table~\ref{tab:cmp_baselines}, we compare our method against the baselines and see a similar trend. SDEdit and DDIM + word swap struggle to retain the structure and the background details. On the other hand, Prompt-to-prompt gets better structure preservation and background error than SDEdit or DDIM + word swap but has a lower CLIP-Acc, indicating that the edit is applied successfully in fewer instances. Our approach gets a high CLIP-Acc while having low Structure Dist and BG LPIPS, showing we can perform the best edit while still retaining the structure and background of the original input image. We show more comparisons of synthetic images in Appendix Figure~\ref{fig:sup_compare_baselines_synthetic}.

\subsection{Ablation Study} \label{sec:ablations}
Finally, we ablate each component of our method and show its effectiveness. Table~\ref{tab:cmp_all_ablations} compares five different configurations. First, config A uses a stochastic noising process for inversion and subsequently swaps the source word with the target edit word (e.g., swapping the word ``cat" with the word ``dog" for the cat $\rightarrow$ dog task). Owing to the stochastic inversion, config A does not retain structure or background from the input and has a high Structure Distance and background error (BG LPIPS). Next, config B replaces the stochastic DDPM inversion with deterministic DDIM inversion and improves both the structure preservation and the background reconstruction. Config C adds the autocorrelation regularization when performing the DDIM inversion, and config D replaces the word swapping with our sentence-based edit directions. Both of these changes cause the desired edit to get applied more consistently, reflected by the improvement in CLIP Acc.
Finally, config E adds the cross-attention guidance $\mathcal{L}_{\text{xa}}$ and corresponds to our final proposed method. The cross-attention guidance helps preserve the structure of the input image and improves both the Structure Dist and BG LPIPS. Figure~\ref{fig:xa_comp} shows this effect of cross-attention guidance qualitatively by comparing config D and config E. When cross-attention guidance is removed, the edited image does not adhere to the input image's structure. E.g. for the task of changing cats to dogs in Figure~\ref{fig:xa_comp}, when the guidance is not used, the edited image contains a dog but in a completely different pose and different background.

\subsection{Model Acceleration with Conditional GANs} \label{sec:comod_distil}
One of the shortcomings of diffusion-based methods is that both inversion and sampling require many steps. To circumvent this and to train a \emph{fast} image-to-image translation model, we can generate a \emph{paired} dataset of input and edited images and train a paired image-conditional GAN that performs a similar edit. %
Figure~\ref{fig:distillation} shows the results obtained by distilling using Co-Mod-GAN~\cite{zhao2021comodgan}. On a NVIDIA A100 GPU with PyTorch, the distilled model only takes 0.018 seconds per image, reducing inference time by a factor of $\sim$ 3,800 times. The distilled conditional GAN can enable real-time applications, while our diffusion-based model can provide high-quality paired training data, which is expensive or impossible to collect manually.

\begin{figure}[t!]
    \centering
    \includegraphics[width=1.0\linewidth]{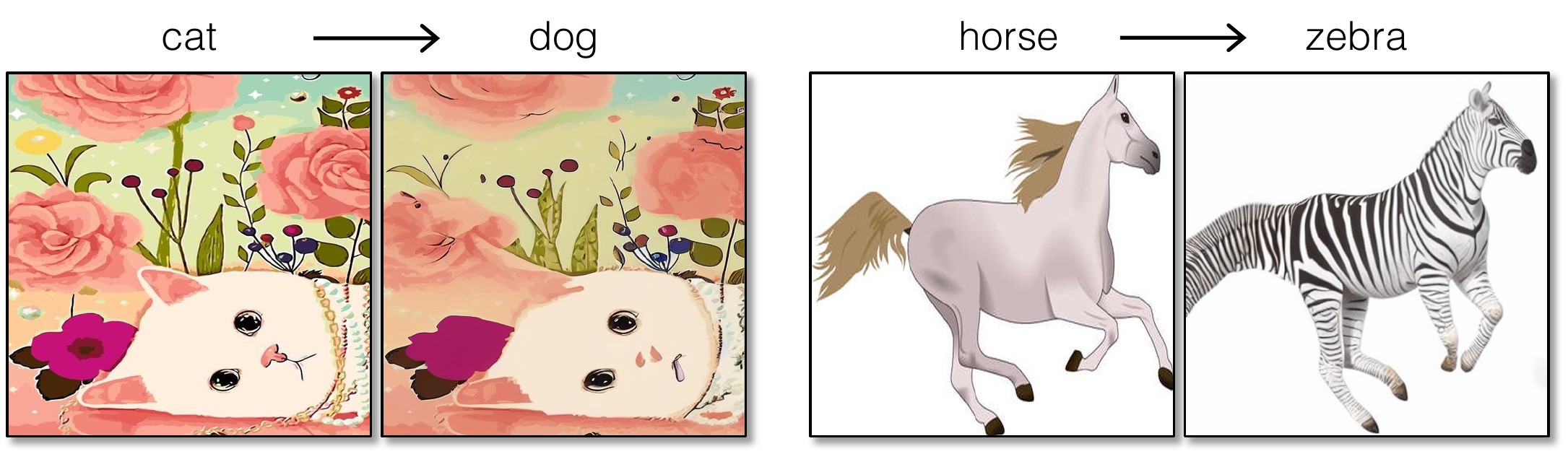}
    \caption{
    \textbf{Limitations.} Our method fails for difficult cases when the object pose is atypical (e.g., the cat on the left) and sometimes for preserving fine spatial position details because of the low resolution of the cross-attention maps (e.g., the leg position and the tail on the right). 
    }
    \label{fig:limitations}
\end{figure}

\section{Limitations and Discussion} \label{sec:conclusion}
We proposed an image-to-image translation method to perform structure-preserving image editing using a pre-trained text-to-image diffusion model. We introduced an automatic way to learn edit direction in the text embedding space. We also proposed cross-attention map guidance to preserve the structure of the original image after applying the learned edit direction. %
We provided detailed quantitative and qualitative results to show the effectiveness of our approach. Our method is training-free and prompting-free. 

\myparagraph{Limitations.}
One of the limitations of our work is that our structure guidance is limited by the resolution of the cross-attention map. For the Stable Diffusion, the resolution for the cross-attention map is $64 \times 64$ which may not be sufficient for very fine-grained structure control (as shown in Figure~\ref{fig:limitations}, our edited zebra does not follow fine-grained details of leg and tail). Although our approach can work with any resolution of cross-attention map, if the base model has a higher resolution for cross-attention map, then our approach can provide even finer structure guidance control. Also, the method can fail in difficult cases of objects having atypical poses (cat in Figure~\ref{fig:limitations}).

\myparagraph{Acknowledgments.} 
This work was partly done by Gaurav Parmar during the Adobe internship.
We thank Sheng-Yu Wang, Gautam Gare, Nupur Kumari, Muyang Li, Ruihan Gao, Aniruddha Mahapatra, and Yotam Nitzan for proofreading our manuscript and feedback. We are also grateful to Kangle Deng, George Cazenavette, Chonghyuk (Andrew) Song, Alyosha Efros, and Phillip Isola for fruitful discussions. This project is partly supported by Adobe Inc.

\typeout{}
{\small
\bibliographystyle{ieee_fullname}
\bibliography{egbib}
}

\clearpage
\appendix
\noindent{\Large\bf Appendix}
\vspace{5pt}

\begin{figure*}[t!]
    \centering
    \includegraphics[width=\linewidth]{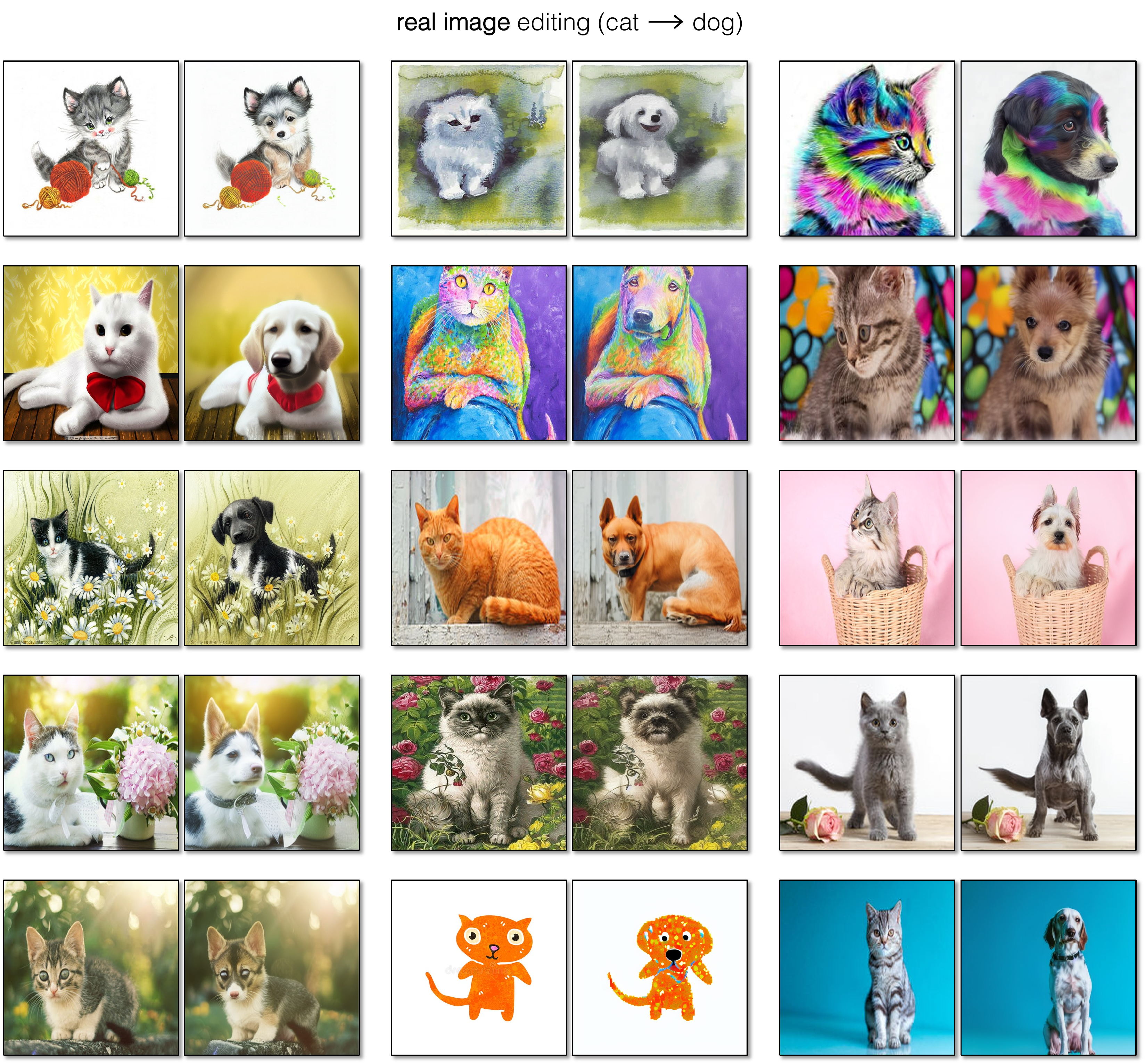}
    \caption{
    \textbf{Additional results.} We show additional results on real images for the cat $\rightarrow$ dog task. }
    \label{fig:sup_grid_results_cat_dog}
\end{figure*}

\begin{figure*}[t!]
    \centering
    \includegraphics[width=\linewidth]{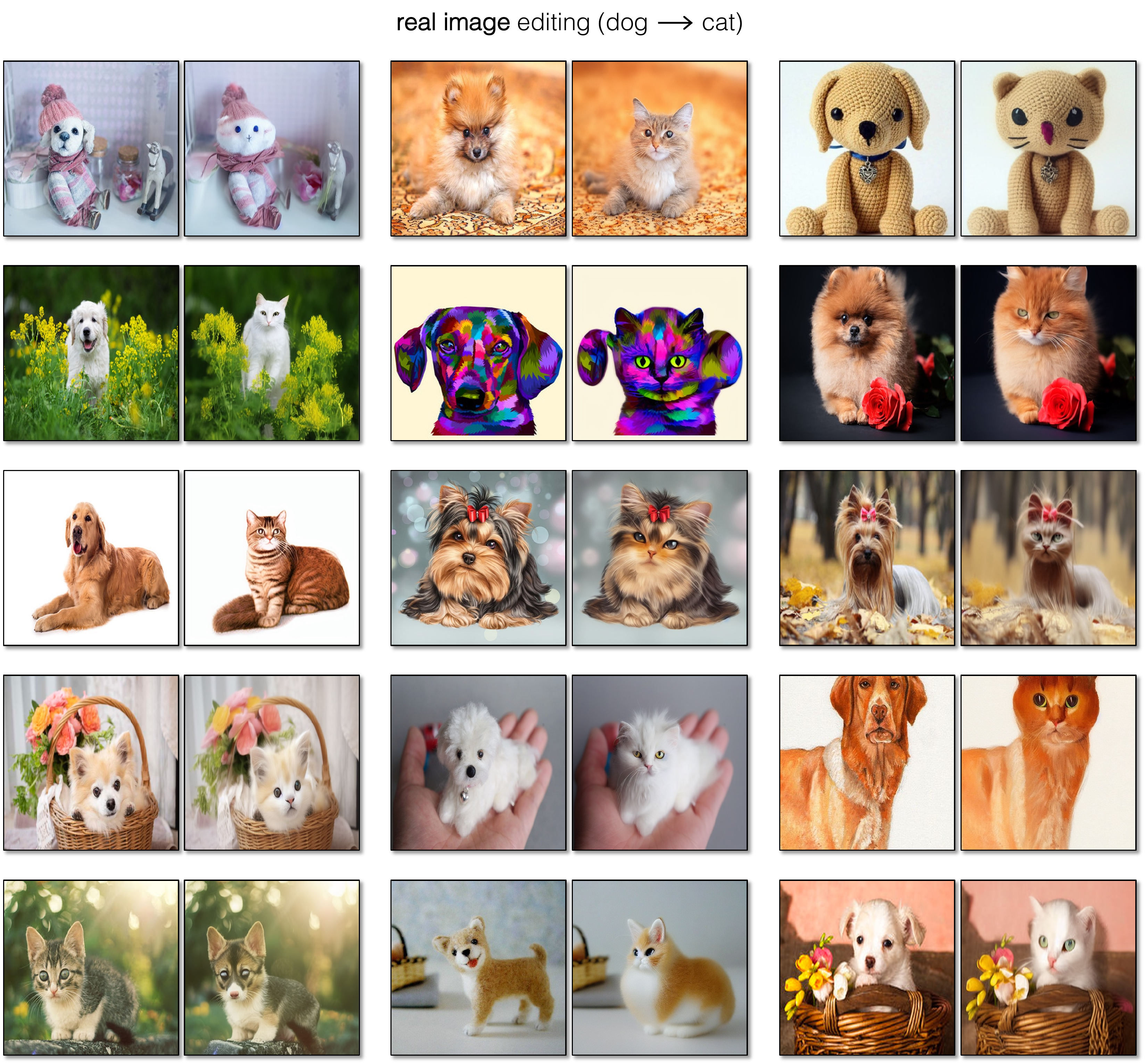}
    \caption{
    \textbf{Additional results.} We show additional results on real images for the dog $\rightarrow$ cat task. }
    \label{fig:sup_grid_results_dog_cat}
\end{figure*}

\begin{figure*}[t!]
    \centering
    \includegraphics[width=\linewidth]{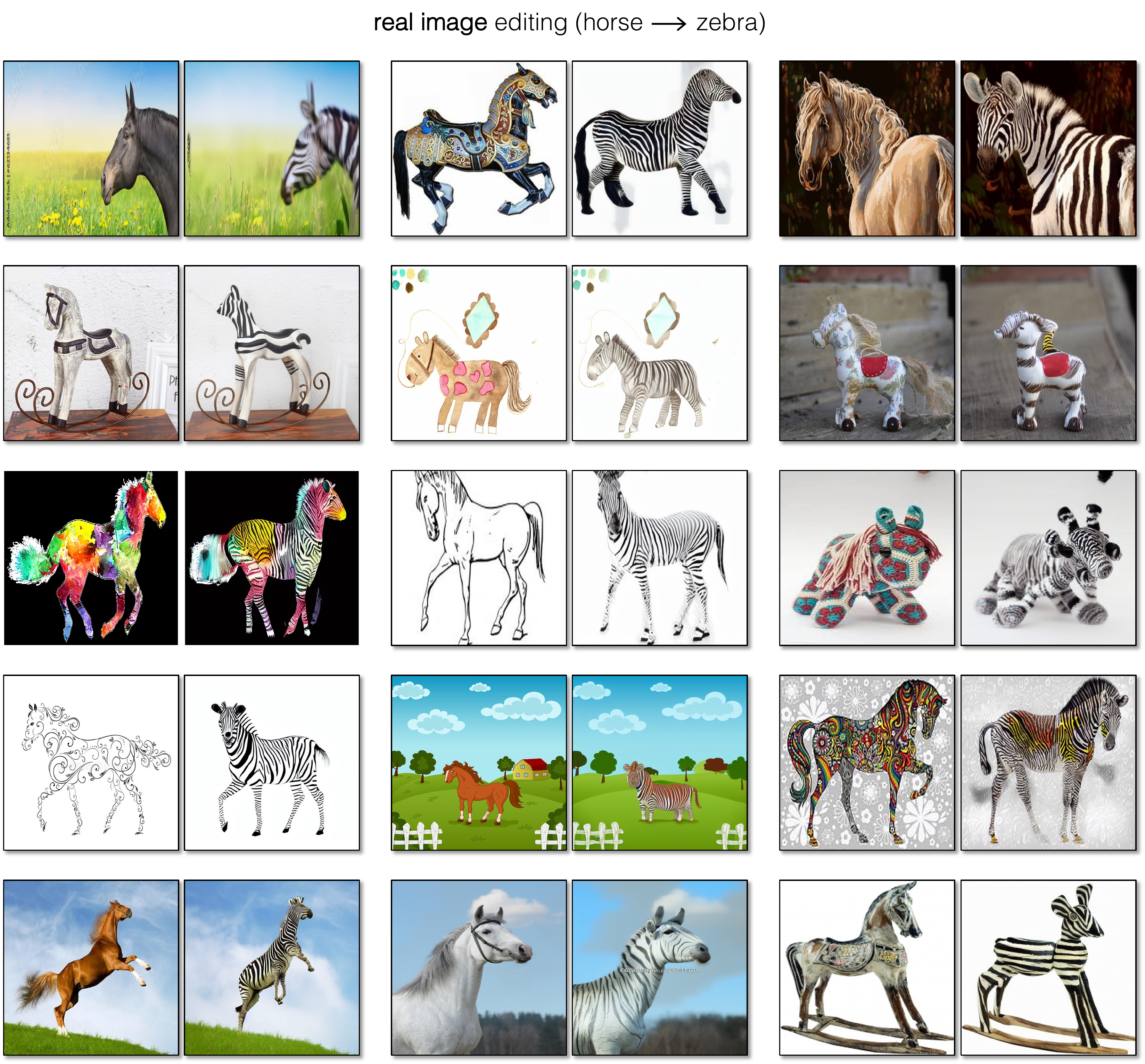}
    \caption{
    \textbf{Additional results.} We show additional results on real images for the horse $\rightarrow$ zebra task. }
    \label{fig:sup_grid_results_horse_zebra}
\end{figure*}

\begin{figure*}[t!]
    \centering
    \includegraphics[width=\linewidth]{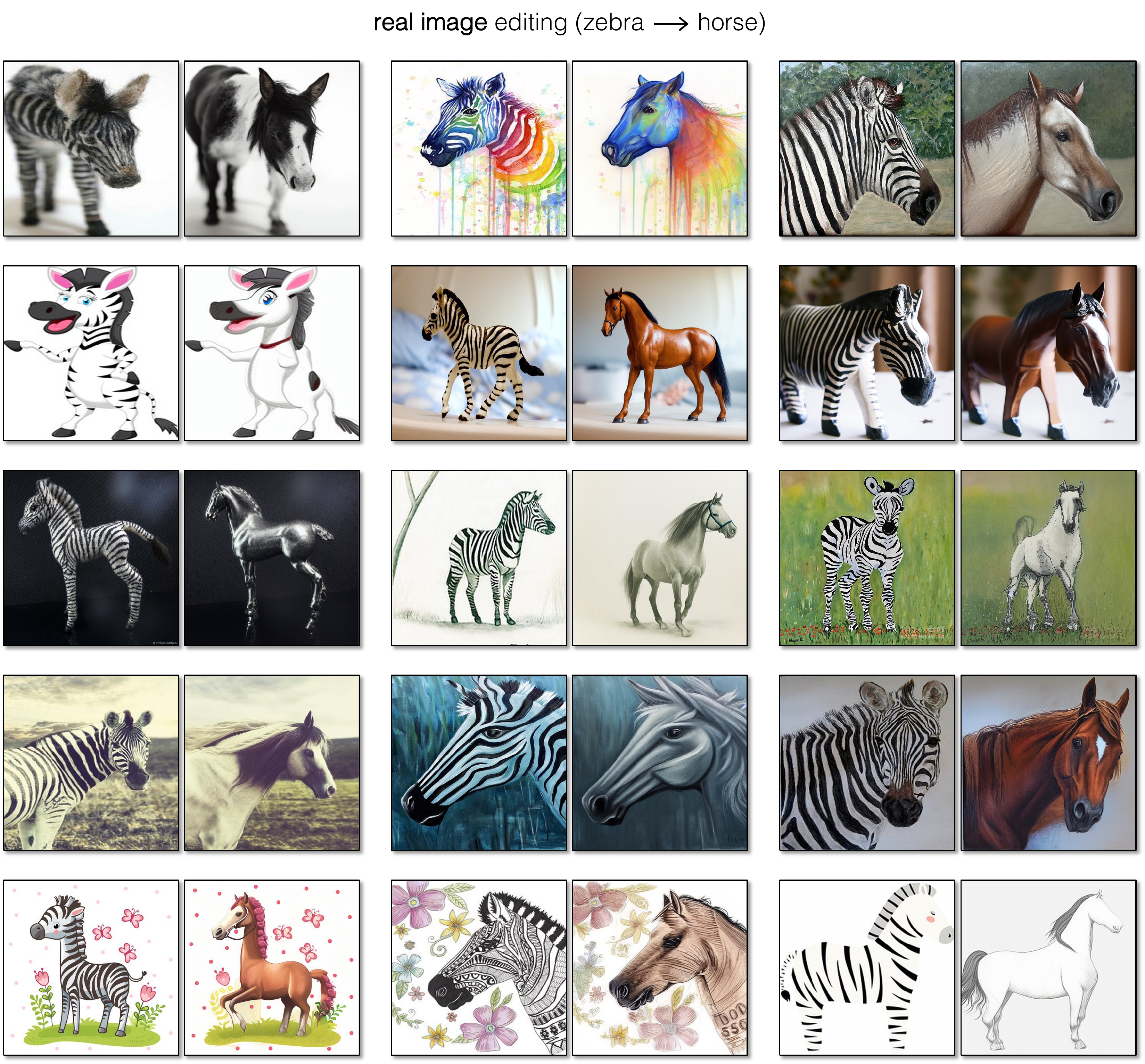}
    \caption{
    \textbf{Additional results.} We show additional results on real images for the zebra $\rightarrow$ horse task. }
    \label{fig:sup_grid_results_zebra_horse}
\end{figure*}

We start with Appendix~\ref{sec:supp_distillation}, which shows additional details about how the fast distilled GAN-based model is trained and provides additional results. 
Next, in Appendix~\ref{sec:supp_base_comp} and Appendix~\ref{sec:supp_ablations}, we show some more comparisons with baselines and the effects of regularization, respectively. 
Appendix~\ref{sec:supp_details} provides the experiment details.
Finally, in Appendix~\ref{sec:supp_limitations}, we discuss some of the societal impacts of this line of research.
We show additional qualitative results in Figures~\ref{fig:sup_grid_results_cat_dog}, ~\ref{fig:sup_grid_results_dog_cat}, ~\ref{fig:sup_grid_results_horse_zebra}, and ~\ref{fig:sup_grid_results_zebra_horse}.

\section{Fast Distillation} \label{sec:supp_distillation}

Section ~\ref{sec:comod_distil} of the main paper discusses distilling a slow, text-to-image diffusion model into a fast, feed-forward model. Here, we describe additional implementation details.

\myparagraph{Paired Dataset Collection.}
We first collect 15,000 pairs of input and edited images generated by our editing method proposed in the main paper. 
Next, we automatically filter out pairs with low segmentation overlap or do not sufficiently increase the CLIP similarity with the target description. For the cat to dog task, we use a segmentation threshold of 0.70 and a CLIP increase threshold of 0.10. For the tree to winter trees and fall trees tasks, we only use a CLIP increase threshold of 0.1 as the off-the-shelf segmentation model does not reliably segment trees in the image. 

\myparagraph{Fast GAN Training.}
Given pairs of input and edited images, we train a CoModGAN~\cite{zhao2021comodgan} to perform image translation. For all experiments, we use a learning rate of 0.001 and a batch size of 64. Additionally, we apply data augmentation in the form of standard color transformations (brightness, contrast, hue, saturation), adding noise, and random crops. We optimize a reconstruction objective using a combination of L1 distance and VGG-based LPIPS~\cite{zhang2018unreasonable}. %

\begin{figure*}[t!]
    \centering
    \includegraphics[width=\linewidth]{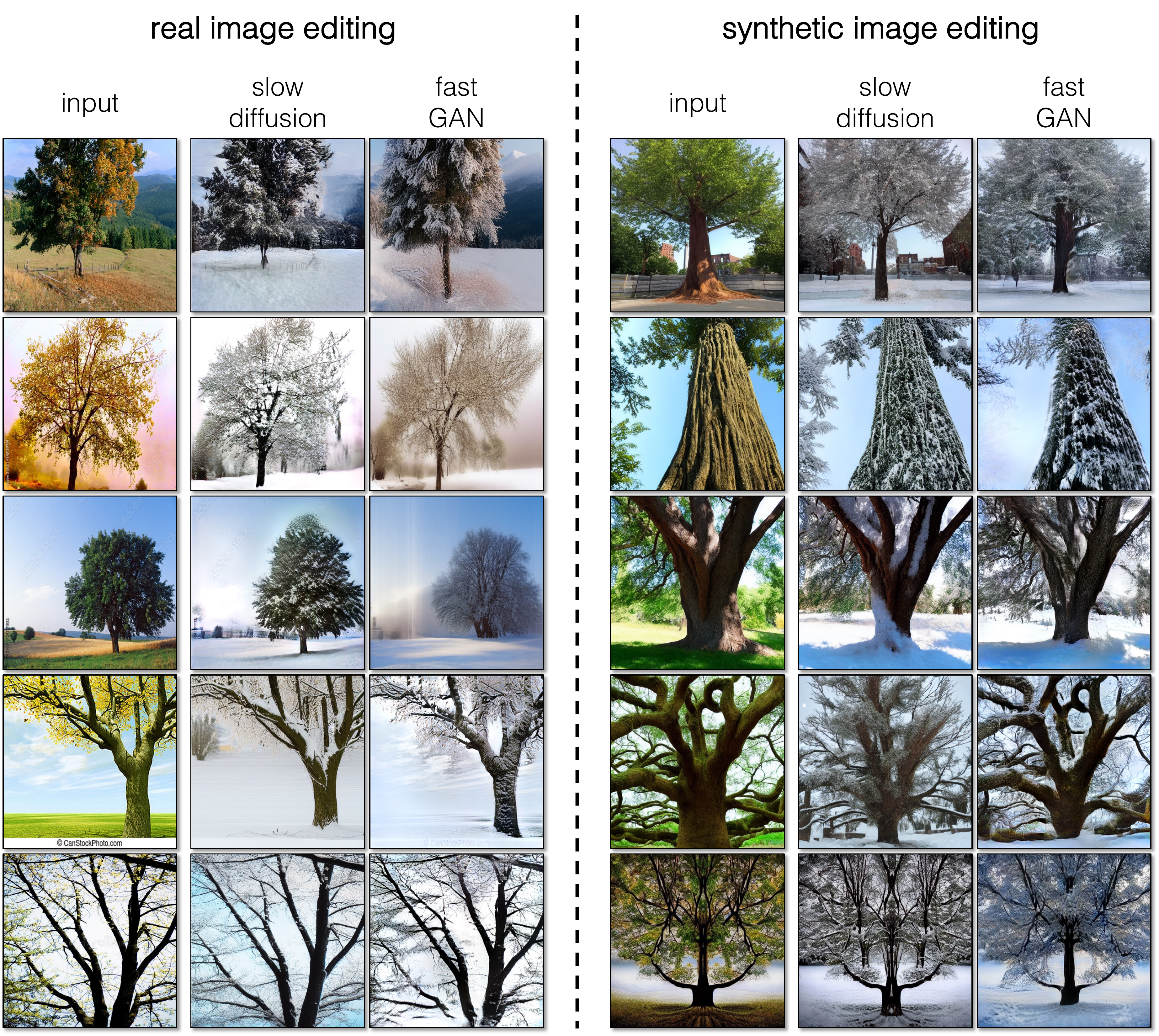}
    \caption{
    \textbf{Model acceleration with conditional GANs.} Here, we show fast GAN distillation and the slower diffusion editing results for the task of tree $\rightarrow$ tree during winter. Our distilled conditional GAN achieves comparable results regarding image quality and structure preservation at a significantly reduced cost.  }
    \label{fig:sup_distillation_tree2winter}
\end{figure*}

\begin{figure*}[t!]
    \centering
    \includegraphics[width=\linewidth]{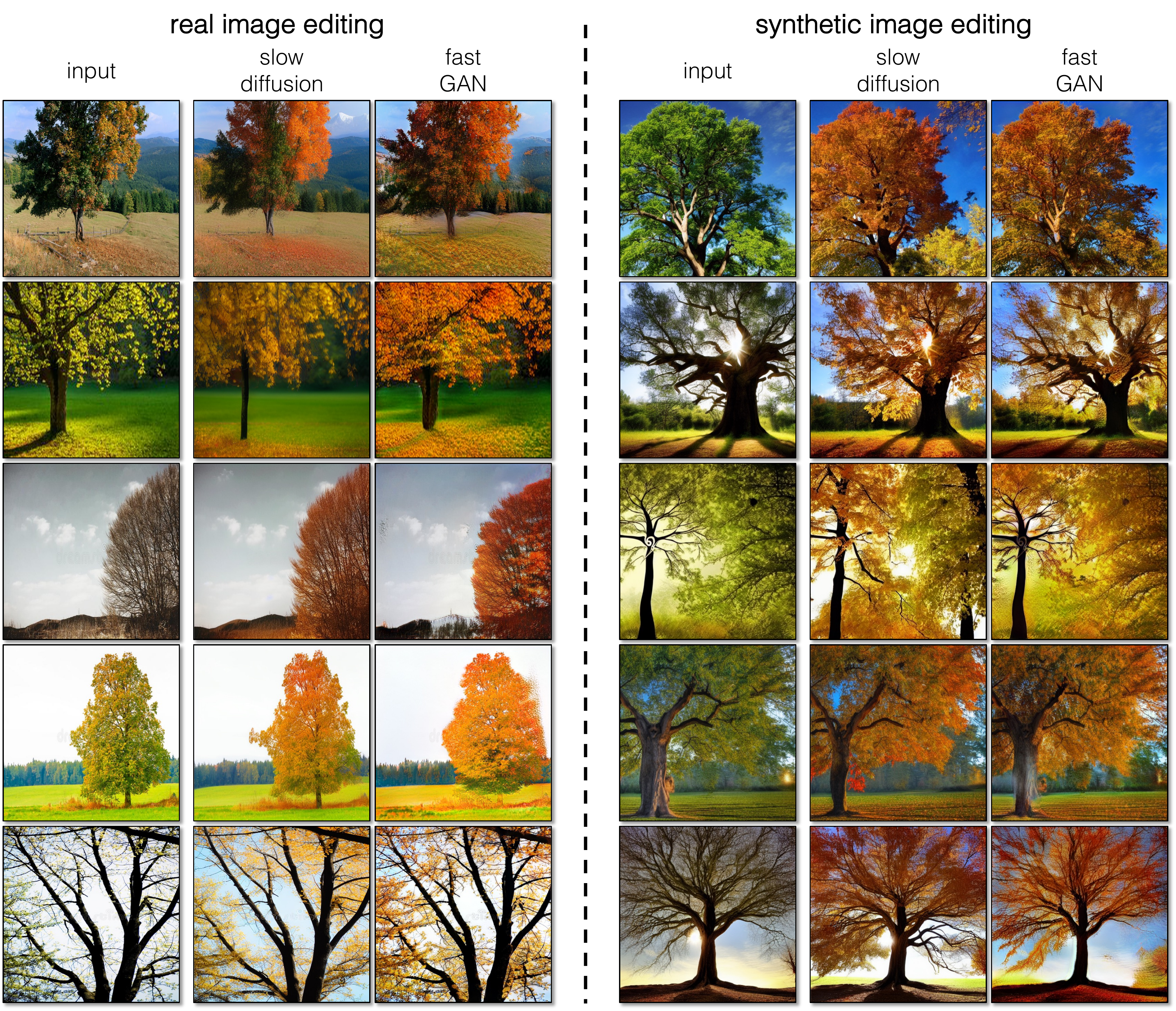}
    \caption{
    \textbf{Model acceleration with conditional GANs.} Here, we show the fast GAN distillation and the slower diffusion editing results for the task of tree $\rightarrow$ tree during fall. Our distilled conditional GAN achieves comparable results regarding image quality and structure preservation at a significantly reduced cost. }
    \label{fig:sup_distillation_tree2fall}
\end{figure*}

\myparagraph{More Results.} 
In Figure~\ref{fig:sup_distillation_tree2winter} and Figure~\ref{fig:sup_distillation_tree2fall}, we show the results of our fast distilled GAN model for the tree to winter tree and fall tree tasks, respectively. Our fast GAN model gives comparable results regarding edit quality and structure perseverance at a much faster inference speed. 
 \begin{figure}[h]
    \centering
    \includegraphics[width=\linewidth]{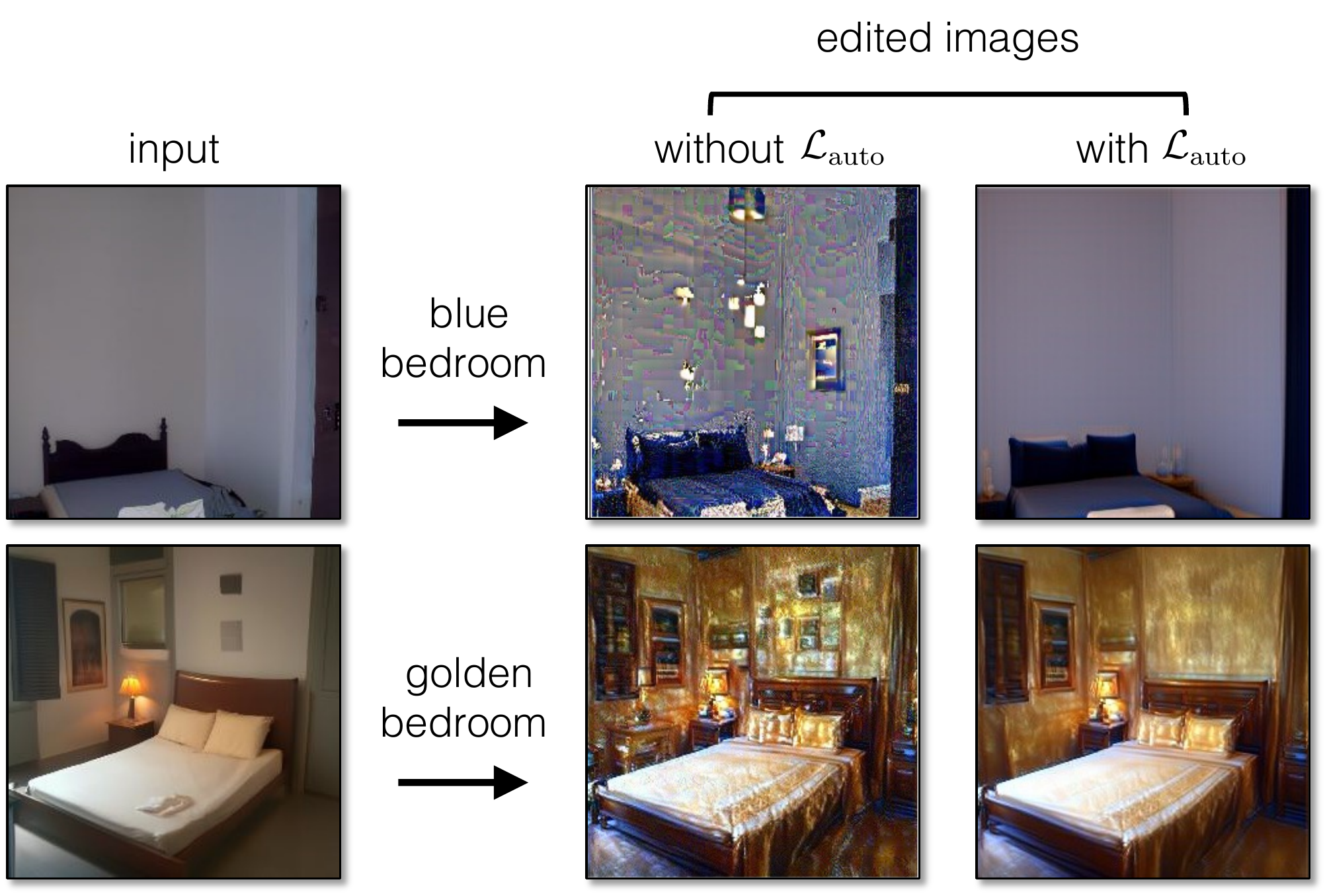}
    \caption{
    \textbf{Qualitative effects of regularization on smaller models.} Here, we show editing results using DiffusionCLIP with and without our regularization. We can see that our regularization is critical for reducing the artifact in the edited results.}  
    \label{fig:regularization_small_models}
\end{figure}


\section{Comparisons to Baselines.} \label{sec:supp_base_comp}

 \begin{figure*}[t!]
    \centering
    \includegraphics[width=\linewidth]{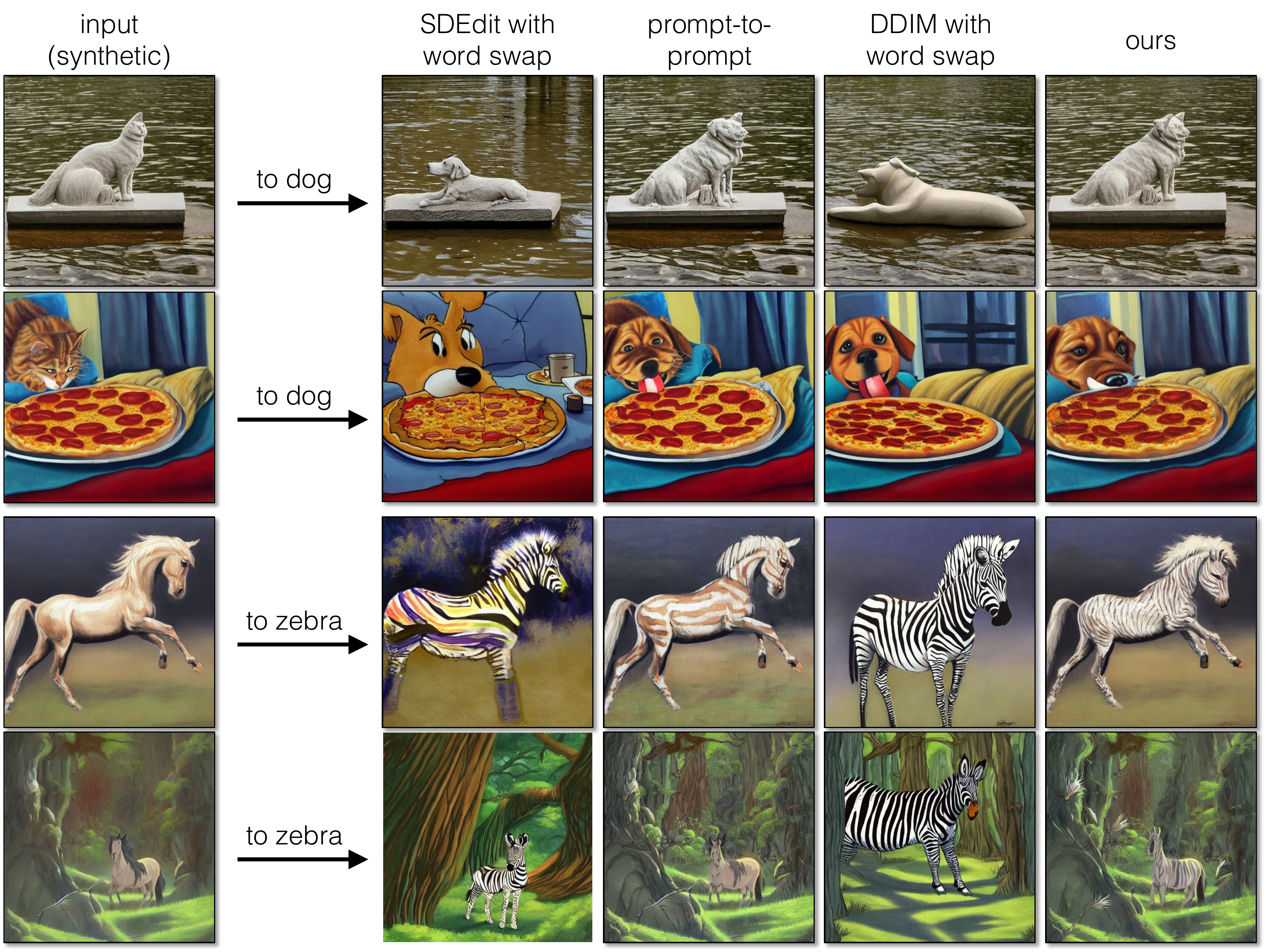}
    \caption{
    \textbf{Comparing to baseline approaches.} We compare our approach with baselines on synthetic images. Our approach does a better job of preserving the structure while performing edits compared to SDEdit~\cite{meng2021sdedit} and DDIM~\cite{song2021ddim} with word swap. prompt-to-prompt~\cite{hertz2022prompt} succeeds to preserve the structure but with lower editing quality.} 
    \label{fig:sup_compare_baselines_synthetic}
\end{figure*}

Figure~\ref{fig:cmp_baselines} and Section~\ref{sec:comparison} in the main paper compare the image editing performance of various methods on \textit{real} images. In Figure~\ref{fig:sup_compare_baselines_synthetic}, we show a similar comparison of synthetic image editing. Our observations are consistent with the real images shown in the main paper. Our method is able to respect the structure of the input image while performing the requested edit. SDEdit~\cite{meng2021sdedit} and DDIM~\cite{song2021ddim} with word swap struggle to preserve the structure. prompt-to-prompt~\cite{hertz2022prompt} works better on synthetic images compared to real images but still struggles to achieve desired edits in some cases (e.g. zebra stripes are not applied correctly).

\section{Ablations} \label{sec:supp_ablations}
\myparagraph{Effects of Regularization during Inversion.}
In Table~\ref{tab:cmp_all_ablations} of the main paper, we show the importance of our regularization $\mathcal{L}_{\text{auto}}$, which was introduced in Section~\ref{sec:inversion} of the main paper. Using this regularization helps improve the extent of editing applied, as indicated by a better CLIP Acc score. 
Our regularization encourages the inverted noise to be more Gaussian, which makes our edit direction more compatible and less inclined to make undesired structure changes. 
We also observe that the effects of the regularization are more pronounced when using smaller-scale diffusion models trained for specific categories. In Figure~\ref{fig:regularization_small_models}, we show image editing results using a smaller diffusion model \cite{dhariwal2021diffusion} trained on the LSUN Bedrooms and finetuned following DiffusionCLIP \cite{Kim_2022_CVPR} to perform the edit. Inverting without regularization and subsequently editing results in noticeable artifacts. 

\section{Experiment Details} \label{sec:supp_details}

\myparagraph{Dataset.}
We use subsets of the LAION 5B dataset for all real image editing experiments. We retrieve 250 relevant images from the dataset by matching CLIP embeddings of the source text description and applying an aesthetics filter of 9 \cite{clip-retrial}. For example, in the cat$\rightarrow$dog translation, we retrieve images from the dataset with a high CLIP similarity with the source word cat. 

\myparagraph{Baselines.}
For all results shown in Figure~\ref{fig:cmp_baselines}, Table ~\ref{tab:cmp_baselines} in the main paper, and Figure~\ref{fig:sup_compare_baselines_synthetic}, we use the official code released by the authors and follow the recommended hyper-parameters.

\myparagraph{Implementation Details.}
For all results shown for our method, we use 100 steps for DDIM inversion and 100 steps for both reconstruction and editing. During DDIM inversion, we apply the noise regularization for 5 iterations at each timestep and use a weight $\lambda$ of 20. Additionally, we use classifier-free guidance \cite{ho2022classifier} for all editing results.

\section{Societal Impact} \label{sec:supp_limitations}

Our work is part of a broader movement toward democratizing content creation with generative models. We aim to allow users to create new content with precise control over the desired edit. Even though the primary usage of our work is in the creative industry, it can be potentially used to fabricate images for malicious practices. However, a line of work has studied whether generated images are detectable, in the context of GANs~\cite{yu2019attributing,nataraj-ei2019,chai2020makes,wang-cvpr2020} and more recently, diffusion models~\cite{corvi2022detection,sha2022fake}. Such work has suggested that while generators produce realistic images, they can still generate consistently detectable artifacts across methods~\cite{chai2020makes,wang-cvpr2020}, enabling their downstream identification.

\end{document}